\PassOptionsToPackage{table}{xcolor}

\documentclass[manuscript,screen,oneside]{acmart}

\usepackage{tabularx}
\usepackage{colortbl}
\usepackage{pifont}
\usepackage{tikz}
\usepackage{graphicx}
\usepackage{forest}
\usepackage{adjustbox}

\usetikzlibrary{calc,positioning,decorations.pathreplacing,arrows.meta,matrix,shadows.blur,shapes.geometric}

\usepackage{booktabs}
\usepackage[table]{xcolor}
\usepackage{colortbl}
\usepackage{pifont}
\usepackage{array}

\definecolor{TableBg}{RGB}{253,252,248} 

\newcolumntype{C}[1]{>{\centering\arraybackslash}p{#1}}

\newcommand{\rcite}[1]{[\textcolor{PaperCiteRed}{\citenum{#1}}]}
\newcommand{\cmark}{\textcolor{green!60!black}{\ding{51}}}
\newcommand{\xmark}{\textcolor{red!70!black}{\ding{55}}}

\renewcommand\footnotetextcopyrightpermission[1]{}
\AtBeginDocument{%
  }

\acmJournal{JACM}

\begin{document}

\title{Agentic Tool Use in Large Language Models: A Survey}

\author{Hu Jinchao}
\affiliation{%
  \institution{School of Computer Science and Technology, Harbin Institute of Technology, Shenzhen}
  \city{Shenzhen}
  \country{China}
}
\email{jchu@stu.hit.edu.cn}

\author{Meizhi Zhong}
\affiliation{%
  \institution{Independent Researcher}
  \city{Beijing}
  \country{China}
}
\email{meizhi.zhong.1999@gmail.com}

\author{Kehai Chen(corresponding author)}
\affiliation{%
  \institution{School of Computer Science and Technology, Harbin Institute of Technology, Shenzhen}
  \city{Shenzhen}
  \country{China}
}
\email{chenkehai@hit.edu.cn}

\author{Xuefeng Bai}
\affiliation{%
  \institution{School of Computer Science and Technology, Harbin Institute of Technology, Shenzhen}
  \city{Shenzhen}
  \country{China}
}
\email{baixuefeng@hit.edu.cn}

\author{Min Zhang}
\affiliation{%
  \institution{School of Computer Science and Technology, Harbin Institute of Technology, Shenzhen}
  \city{Shenzhen}
  \country{China}
}
\email{zhangmin2021@hit.edu.cn}

\renewcommand{\shortauthors}{Hu Jinchao}

\begin{abstract}
  Large language models are increasingly being deployed as autonomous agents yet their real world effectiveness depends on reliable tools for information retrieval, computation and external action. Existing studies remain fragmented across tasks, tool types, and training settings, lacking a unified view of how tool-use methods differ and evolve. This survey organizes the literature into three paradigms: prompting as plug-and-play, supervised tool learning and reward-driven tool policy learning, analyzes their methods, strengths and failure modes, reviews the evaluation landscape and highlights key challenges, aiming to address this fragmentation and provide a more structured evolutionary view of agentic tool use.
\end{abstract}

\ccsdesc[500]{Computing methodologies~Natural language generation}

\keywords{Large language models,Agentic tool use,Prompt engineering,Instruction tuning,Reinforcement learning}


\maketitle

\section{Introduction}

Artificial intelligence (AI) has evolved from early rule-based expert systems to modern data-driven machine learning~\cite{Bishop2006PRML}. The Transformer architecture~\cite{Vaswani2017Attention} catalyzed this transition and enabled the rise of large language models (LLMs). Foundational models such as BERT~\cite{Devlin2019BERT} and T5~\cite{Raffel2020T5} established strong paradigms for natural language understanding and generation. Subsequent models, including Llama 2~\cite{Touvron2023Llama2}, Qwen~\cite{Qwen2025Qwen25}, Mixtral~\cite{Jiang2024Mixtral}, and GPT-4~\cite{OpenAI2023GPT4}, have demonstrated unprecedented capability. Pretrained on massive corpora~\cite{Gao2020Pile}, these models exhibit emergent strengths~\cite{Wei2022EmergentAbilities} in contextual understanding, complex reasoning~\cite{Kojima2022ZeroShotReasoners}, and content generation.

However, standard LLMs remain inherently static. Their knowledge is bounded by training data, which can lead to factual errors or "hallucinations"~\cite{Lewis2020RAG}. They also struggle to access real-time information, execute precise computation, or interact reliably with external systems~\cite{Ji2023HallucinationSurvey}. These limitations, together with safety and bias risks inherited from web-scale data~\cite{Bender2021StochasticParrots}, constrain deployment in dynamic real-world settings.

To overcome these boundaries, agentic AI has emerged as a central paradigm~\cite{Zhao2023LLMAIAgentsSurvey,Wang2024LLMAutonomousAgents,Mialon2023AugmentedLMs}. In this paradigm, the LLM serves as a reasoning core but is augmented with planning, memory, and action capabilities. Tool use is the key mechanism that operationalizes action~\cite{Schick2023Toolformer,Shen2023HuggingGPT}. By invoking external tools (e.g., APIs, databases, and code interpreters), an LLM agent can access up-to-date knowledge through search~\cite{Nakano2021WebGPT}, improve computational accuracy through programmatic execution~\cite{Gao2023PAL}, and perform real-world operations through external systems~\cite{Shen2023HuggingGPT}.

The growing importance of tool use has driven substantial research activity, including multiple surveys that attempt to structure the field. Existing reviews cover complementary perspectives: some emphasize agent architecture (planning, memory, and action)~\cite{Zhao2023LLMAIAgentsSurvey}; others review augmented language models and external memory/tool mechanisms~\cite{Mialon2023AugmentedLMs,Zhang2025MemoryAgents}; and still others focus on tool taxonomies and data construction pipelines~\cite{Qu2025ToolLearningLLM}. Despite their value, many of these frameworks provide a largely static snapshot. What remains insufficiently articulated is the methodological evolution from prompting as plug-and-play to supervised tool learning and reward-driven policy learning. This missing evolutionary view can obscure both continuity and trade-offs across paradigms. Table~\ref{tab:survey_comparison} compares representative surveys along six dimensions: chronological evolution, paradigm taxonomy, tool coverage, training data, evaluation landscape, and open challenges.

\begin{table}[t]
  \centering
  \footnotesize
  \setlength{\tabcolsep}{13.0pt}
  \renewcommand{\arraystretch}{1.5}
  \caption{Comparison of representative survey papers on agentic tool use and function calling.}
  \label{tab:survey_comparison}

  \begin{tabular}{lcccccc}
    \toprule
    \textbf{Survey Paper} &
    \textbf{\shortstack{Chronological\\ evolution}} &
    \textbf{\shortstack{Paradigm\\ taxonomy}} &
    \textbf{\shortstack{Tool\\ coverage}} &
    \textbf{\shortstack{Training\\ data}} &
    \textbf{\shortstack{Evaluation\\ landscape}} &
    \textbf{\shortstack{Open\\ challenges}} \\
    \midrule
    Zhao et al.~\cite{Zhao2023LLMAIAgentsSurvey}                          & \cmark & \xmark & \xmark & \xmark & \cmark & \cmark \\
    Wang et al.~\cite{Wang2024LLMAutonomousAgents}                        & \cmark & \xmark & \xmark & \xmark & \cmark & \cmark \\
    Mialon et al.~\cite{Mialon2023AugmentedLMs}                           & \cmark & \xmark & \xmark & \xmark & \cmark & \cmark \\
    Qu et al.~\cite{Qu2025ToolLearningLLM}                             & \xmark & \cmark & \cmark & \cmark & \cmark & \cmark \\
    Shen~\cite{Shen2024LLMWithToolsSurvey}                                & \xmark & \cmark & \cmark & \cmark & \xmark & \cmark \\
    Chen et al.~\cite{Chen2025ToolLearningComprehensiveSurvey}            & \xmark & \cmark & \cmark & \cmark & \cmark & \cmark \\
    \textbf{Ours}                                                        & \cmark & \cmark & \cmark & \cmark & \cmark & \cmark \\
    \bottomrule
  \end{tabular}%
\end{table} 

In this paper, we make several key contributions to the field of LLM-based agentic tool use:
\begin{itemize}
  \item We provide an evolutionary synthesis of agentic tool use, tracing how the field progresses from prompting-based control to supervised tool learning and reward-driven policy learning.
  \item We propose a unified three-paradigm taxonomy with explicit methodological boundaries, clarifying the role of optimization signals (in-context control, supervised signals, and reward feedback).
  \item We systematize the evaluation landscape as a cross-cutting dimension, connecting method families to benchmark levels from function-call correctness to end-to-end interactive success, safety, and robustness.
  \item We identify unresolved challenges and practical research directions, including long-horizon credit assignment, scalable tool generalization, and alignment-aware deployment.
\end{itemize}

In this survey, we synthesize agentic tool use from an explicitly evolutionary perspective. We organize prior work into three methodological paradigms: (I) Prompting as Plug-and-Play, where a frozen model is guided to use tools through prompts and in-context interaction instead of weight updates~\cite{Xie2022ICLBayes}; (II) Supervised Tool Learning, which uses labeled or synthetic supervision to internalize robust tool-use behavior~\cite{Shen2023HuggingGPT}; and (III) Reward-Driven Tool Policy Learning, which optimizes long-horizon interaction through reinforcement and reward feedback~\cite{Shinn2023Reflexion}. We further analyze evaluation as a cross-cutting dimension and track its evolution from function-call correctness~\cite{Li2023APIBank} to end-to-end task success in interactive settings~\cite{Liu2024AgentBench}. Through this lens, we aim to provide a coherent map of the field, clarify the design trade-offs linking paradigms, and connect foundational methods to emerging autonomous agent systems. Figure~\ref{fig:agent-tool-use-overview} summarizes this framework.

\definecolor{MainNode}{HTML}{5E6F65}

\definecolor{Prompting}{HTML}{5AA0E8}    
\definecolor{Training}{HTML}{F2A84A}      
\definecolor{Policy}{HTML}{A874D6}        
\definecolor{Evaluation}{HTML}{3FCF92}    

\definecolor{SubCategory}{HTML}{7F8C8D}   
\definecolor{PaperLink}{HTML}{000000}     
\definecolor{PaperCiteRed}{HTML}{C0392B}  
\definecolor{PaperBg}{HTML}{E6E6E6}

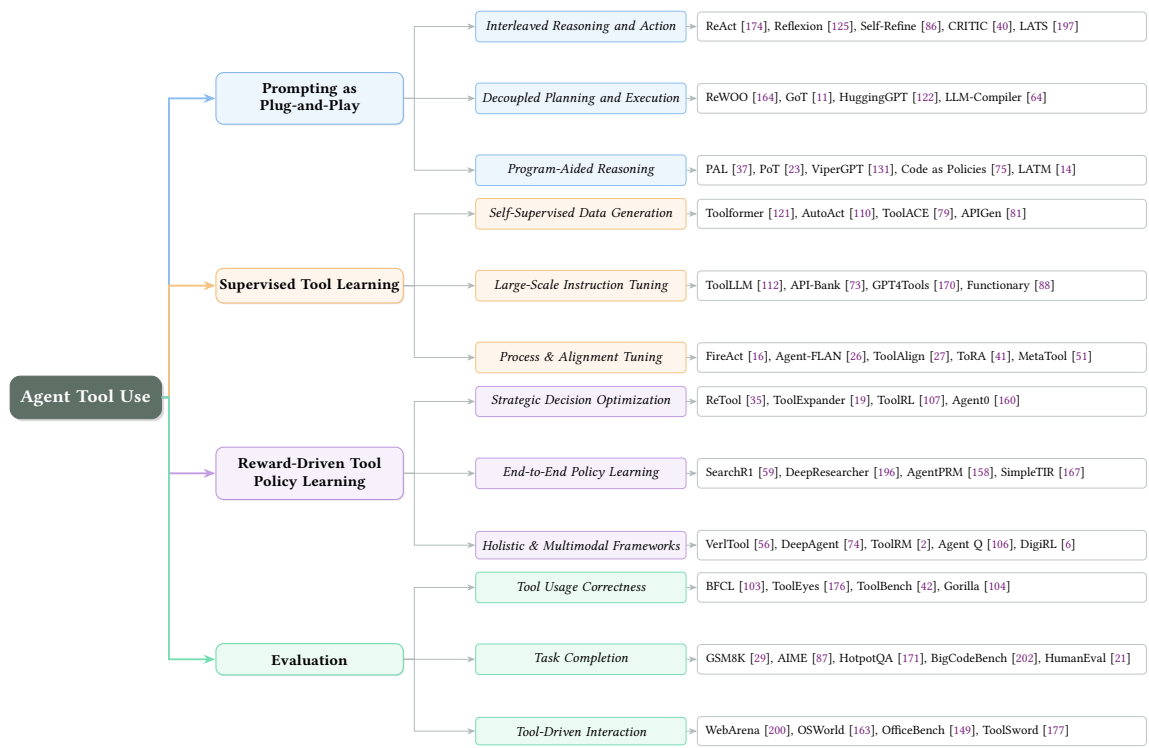
\begin{figure*}[t]
  \centering
  \begin{adjustbox}{max width=\linewidth}
  \begin{tikzpicture}[
    node distance=3.0cm and 4.5cm,
    main/.style={
      rectangle, rounded corners=8pt,
      text width=4.0cm,
      align=center,
      inner ysep=10pt,
      draw=MainNode, fill=MainNode,
      text=white, font=\bfseries\huge,
      drop shadow={opacity=0.3, shadow xshift=2pt, shadow yshift=-2pt}
    },
    paradigm/.style={
      rectangle, rounded corners=6pt,
      text width=5.0cm,
      align=center,
      inner ysep=8pt,
      draw=#1!80, fill=#1!10,
      text=black, font=\bfseries\LARGE,
      drop shadow={opacity=0.2, shadow xshift=1pt, shadow yshift=-1pt}
    },
    subcategory/.style={
      rectangle, rounded corners=3pt,
      text width=5.5cm,
      align=center,
      inner ysep=7pt,
      draw=SubCategory!50, fill=SubCategory!5,
      text=black, font=\Large\itshape,
      inner xsep=5pt
    },
    papers/.style={
      rectangle, rounded corners=3pt,
      text width=12cm,
      draw=SubCategory!50, fill=white,
      text=PaperLink, font=\large,
      align=left, 
      inner sep=7pt
    },
    connector/.style={
      ->, >=Stealth, line width=1.5pt
    },
    subconnector/.style={
      ->, >=Stealth, line width=0.8pt, draw=SubCategory!60
    },
    connectorP/.style={connector, draw=Prompting!70},
    connectorT/.style={connector, draw=Training!70},
    connectorPo/.style={connector, draw=Policy!70},
    connectorE/.style={connector, draw=Evaluation!70}
  ]

  \node[main] (main) {Agent Tool Use};

  \node[paradigm=Prompting, above right=7cm and 1.5cm of main] (prompting) 
    {Prompting as Plug-and-Play};
  \node[paradigm=Training, below=4cm of prompting] (training) 
    {Supervised Tool Learning};
  \node[paradigm=Policy, below=4cm of training] (policy) 
    {Reward-Driven Tool Policy Learning};
  \node[paradigm=Evaluation, below=4cm of policy] (evaluation) 
    {Evaluation};

  \draw[connectorP] (main.east) -- ++(0.2,0) |- (prompting.west);
  \draw[connectorT] (main.east) -- ++(0.2,0) |- (training.west);
  \draw[connectorPo] (main.east) -- ++(0.2,0) |- (policy.west);
  \draw[connectorE] (main.east) -- ++(0.2,0) |- (evaluation.west);

  \node[subcategory, draw=Prompting!70, fill=Prompting!10, right=2.0cm of prompting, yshift=2.0cm] (p1) {Interleaved Reasoning and Action};
  \node[papers,right=0.3cm of p1.east, anchor=west,append after command=
  {\pgfextra
      \draw[subconnector] (p1.east) -- ++(0.25,0) |- (\tikzlastnode.west);
    \endpgfextra}
       ] (p1papers)
    {ReAct \rcite{Yao2023ReAct}, Reflexion \rcite{Shinn2023Reflexion}, Self-Refine \rcite{Madaan2023SelfRefine}, CRITIC \rcite{Gou2024Critic}, LATS \rcite{Zhou2023LATS}};
  
  \node[subcategory, draw=Prompting!70, fill=Prompting!10, right=2.0cm of prompting] (p2) {Decoupled Planning and Execution};
  \node[papers, right=0.3cm of p2.east, anchor=west,
  append after command={
    \pgfextra
      \draw[subconnector] (p2.east) -- ++(0.25,0) |- (\tikzlastnode.west);
    \endpgfextra}
       ] (p2papers)
    {ReWOO \rcite{Xu2023ReWOO}, GoT \rcite{Besta2024GraphOfThoughts}, HuggingGPT \rcite{Shen2023HuggingGPT}, LLM-Compiler \rcite{Kim2024LLMCompiler}};
  
  \node[subcategory, draw=Prompting!70, fill=Prompting!10, right=2.0cm of prompting, yshift=-2.0cm] (p3) {Program-Aided Reasoning};
  \node[papers, right=0.3cm of p3.east, anchor=west,
  append after command={
    \pgfextra
      \draw[subconnector] (p3.east) -- ++(0.25,0) |- (\tikzlastnode.west);
    \endpgfextra}
       ] (p3papers)
    {PAL \rcite{Gao2023PAL}, PoT \rcite{Chen2023ProgramOfThoughts}, ViperGPT \rcite{Suris2023ViperGPT}, Code as Policies \rcite{Liang2022CodeAsPolicies}, LATM \rcite{Cai2024LLMToolMakers}};

  \draw[subconnector] (prompting.east) -- ++(0.3,0) |- (p1.west);
  \draw[subconnector] (prompting.east) -- ++(0.3,0) |- (p2.west);
  \draw[subconnector] (prompting.east) -- ++(0.3,0) |- (p3.west);

  \node[subcategory, draw=Training!70, fill=Training!10, right=2.0cm of training, yshift=2.0cm] (t1) {Self-Supervised Data Generation};
  \node[papers, right=0.3cm of t1.east, anchor=west,
  append after command={
    \pgfextra
      \draw[subconnector] (t1.east) -- ++(0.25,0) |- (\tikzlastnode.west);
    \endpgfextra}
       ] (t1papers)
    {Toolformer \rcite{Schick2023Toolformer}, AutoAct \rcite{Qiao2023AutoAct}, ToolACE \rcite{Liu2024ToolACE}, APIGen \rcite{Liu2024APIGen}};

  \node[subcategory, draw=Training!70, fill=Training!10, right=2.0cm of training] (t2) {Large-Scale Instruction Tuning};
  \node[papers, right=0.3cm of t2.east, anchor=west,
  append after command={
    \pgfextra
      \draw[subconnector] (t2.east) -- ++(0.25,0) |- (\tikzlastnode.west);
    \endpgfextra}
       ] (t2papers)
    {ToolLLM \rcite{Qin2023ToolLLM}, API-Bank \rcite{Li2023APIBank}, GPT4Tools \rcite{Yang2023GPT4Tools}, Functionary \rcite{MeetKai2024Functionary}};

  \node[subcategory, draw=Training!70, fill=Training!10, right=2.0cm of training, yshift=-2.0cm] (t3) {Process \& Alignment Tuning};
  \node[papers, right=0.3cm of t3.east, anchor=west,
  append after command={
    \pgfextra
      \draw[subconnector] (t3.east) -- ++(0.25,0) |- (\tikzlastnode.west);
    \endpgfextra}
       ] (t3papers)
    {FireAct \rcite{Chen2023FireAct}, Agent-FLAN \rcite{Chen2024AgentFLAN}, ToolAlign \rcite{Chen2024ToolAlign}, ToRA \rcite{Gou2024ToRA}, MetaTool \rcite{Huang2024MetaTool}};

  \draw[subconnector] (training.east) -- ++(0.3,0) |- (t1.west);
  \draw[subconnector] (training.east) -- ++(0.3,0) |- (t2.west);
  \draw[subconnector] (training.east) -- ++(0.3,0) |- (t3.west);

   \node[subcategory, draw=Policy!70, fill=Policy!10, right=2.0cm of policy, yshift=2.0cm] (po1) {Strategic Decision Optimization};
  \node[papers, right=0.3cm of po1.east, anchor=west,
  append after command={
    \pgfextra
      \draw[subconnector] (po1.east) -- ++(0.25,0) |- (\tikzlastnode.west);
    \endpgfextra}
       ] (po1papers)
    {ReTool \rcite{Feng2025ReTool}, ToolExpander \rcite{Chen2025ToolExpander}, ToolRL \rcite{Qian2025ToolRL}, Agent0 \rcite{Xia2025Agent0}};
  
  \node[subcategory, draw=Policy!70, fill=Policy!10, right=2.0cm of policy] (po2) {End-to-End Policy Learning};
  \node[papers, right=0.3cm of po2.east, anchor=west,
  append after command={
    \pgfextra
      \draw[subconnector] (po2.east) -- ++(0.25,0) |- (\tikzlastnode.west);
    \endpgfextra}
       ] (po2papers)
    {SearchR1 \rcite{Jin2025SearchR1}, DeepResearcher \rcite{Zheng2025DeepResearcher}, AgentPRM \rcite{Xi2025AgentPRM}, SimpleTIR \rcite{Xue2025SimpleTIR}};
  
  \node[subcategory, draw=Policy!70, fill=Policy!10, right=2.0cm of policy, yshift=-2.0cm] (po3) {Holistic \& Multimodal Frameworks};
  \node[papers, right=0.3cm of po3.east, anchor=west,
  append after command={
    \pgfextra
      \draw[subconnector] (po3.east) -- ++(0.25,0) |- (\tikzlastnode.west);
    \endpgfextra}
       ] (po3papers)
    {VerlTool \rcite{Jiang2025VerlTool}, DeepAgent \rcite{Li2025DeepAgent}, ToolRM \rcite{Agarwal2025ToolRM}, Agent Q \rcite{Putta2024AgentQ}, DigiRL \rcite{Bai2024DigiRL}};

  \draw[subconnector] (policy.east) -- ++(0.3,0) |- (po1.west);
  \draw[subconnector] (policy.east) -- ++(0.3,0) |- (po2.west);
  \draw[subconnector] (policy.east) -- ++(0.3,0) |- (po3.west);

  \node[subcategory, draw=Evaluation!70, fill=Evaluation!10, right=2.0cm of evaluation, yshift=2.0cm] (e1) {Tool Usage Correctness};
  \node[papers, right=0.3cm of e1.east, anchor=west,
  append after command={
    \pgfextra
      \draw[subconnector] (e1.east) -- ++(0.25,0) |- (\tikzlastnode.west);
    \endpgfextra}
       ] (e1papers)
    {BFCL \rcite{Patil2025BFCL}, ToolEyes \rcite{Ye2025ToolEyes}, ToolBench \rcite{Guo2024StableToolBench}, Gorilla \rcite{Patil2024Gorilla}};
  
  \node[subcategory, draw=Evaluation!70, fill=Evaluation!10, right=2.0cm of evaluation] (e2) {Task Completion};
  \node[papers, right=0.3cm of e2.east, anchor=west,
  append after command={
    \pgfextra
      \draw[subconnector] (e2.east) -- ++(0.25,0) |- (\tikzlastnode.west);
    \endpgfextra}
       ] (e2papers)
    {GSM8K \rcite{Cobbe2021GSM8K}, AIME \rcite{MAA_AIME}, HotpotQA \rcite{Yang2018HotpotQA}, BigCodeBench \rcite{Zhuo2024BigCodeBench}, HumanEval \rcite{Chen2021Codex}};
  
  \node[subcategory, draw=Evaluation!70, fill=Evaluation!10, right=2.0cm of evaluation, yshift=-2.0cm] (e3) {Tool-Driven Interaction};
  \node[papers, right=0.3cm of e3.east, anchor=west,
  append after command={
    \pgfextra
      \draw[subconnector] (e3.east) -- ++(0.25,0) |- (\tikzlastnode.west);
    \endpgfextra}
       ] (e3papers)
    {WebArena \rcite{Zhou2024WebArena}, OSWorld \rcite{Xie2024OSWorld}, OfficeBench \rcite{Wang2024OfficeBench}, ToolSword \rcite{Ye2024ToolSword}};

  \draw[subconnector] (evaluation.east) -- ++(0.3,0) |- (e1.west);
  \draw[subconnector] (evaluation.east) -- ++(0.3,0) |- (e2.west);
  \draw[subconnector] (evaluation.east) -- ++(0.3,0) |- (e3.west);

  \end{tikzpicture}
  \end{adjustbox}
  \caption{Overview of the Agentic Tool Use framework.}
  \Description{A comprehensive taxonomy diagram showing the evolution of agentic tool use across four branches: prompting as plug-and-play, supervised tool learning, reward-driven tool policy learning, and evaluation.}
  \label{fig:agent-tool-use-overview}
\end{figure*}

\begin{itemize}
  \item \textbf{Prompting as Plug-and-Play.}
  This initial approach uses frozen LLM parameters and relies on in-context learning and prompt engineering.
  It requires no model training.
  This category explores control flows such as interleaved reasoning and acting cycles that mix thought and action, and decoupled planning and execution that separates planning from action.
  While highly flexible, this method often suffers from high latency and token costs, and its reliability can be brittle for complex, multi step tasks.

  \item \textbf{Supervised Tool Learning.}
  This approach moves from prompt engineering to data engineering to overcome the scaling and reliability bottlenecks of tuning-free methods.
  The core idea is to use supervised fine-tuning (SFT) to teach models the syntax and logic of tool use, internalizing this skill into model parameters.
  The primary challenge is creating large-scale, high-quality datasets.
  Key strategies include self-supervised data generation and teacher-guided instruction synthesis for complex multi-step tool-use trajectories.

  \item \textbf{Reward-Driven Tool Policy Learning.}
  This paradigm shifts from static imitation to dynamic optimization, treating tool use as a sequential decision-making problem.
  It uses reinforcement learning (RL) to train the agent to discover an effective policy in dynamic environments rather than mimicking fixed trajectories.
  The research targets strategic decision making, including when to call a tool versus relying on model knowledge, and multi-tool credit assignment in long tool-call chains.

  \item \textbf{Evaluation.}
  This category summarizes evaluation for tool-using agents and has co-evolved with the methodological paradigms.
  Benchmarks progressed from function-call correctness (valid syntax and arguments) to end-to-end task success in interactive environments.
  Recent benchmarks further evaluate whether agents follow safety rules and operational guidelines while completing tasks.
\end{itemize}
Notably, this survey emphasizes the complementary nature of the three paradigms and argues that production-grade systems will likely combine all of them. We further summarize unresolved challenges and promising research directions, including multi-tool credit assignment in long-horizon tasks, autonomous tool creation and self-extension, and the integration of safety, alignment, and compliance into deployed agent systems.
The organization of the rest of this survey is as follows. We begin with the background of this survey in \autoref{sec:background}, introducing the development timeline of LLM tool use. We then discuss three methodological paradigms: prompting as plug-and-play (\autoref{sec:prompting-plug-and-play}), supervised tool learning (\autoref{sec:training-internalized}), and reward-driven tool policy learning (\autoref{sec:policy-autonomous}). We treat evaluation as a cross-cutting dimension and summarize the science and benchmarks of evaluation in \autoref{sec:evaluation}. In the end, we conclude this paper with a summary of the three complementary paradigms and a discussion about our future outlooks in \autoref{sec:future-directions}.

\section{Background of Agent Tool Use}
\label{sec:background}
\begin{figure}[H]
  \centering
  \includegraphics[width=\columnwidth]{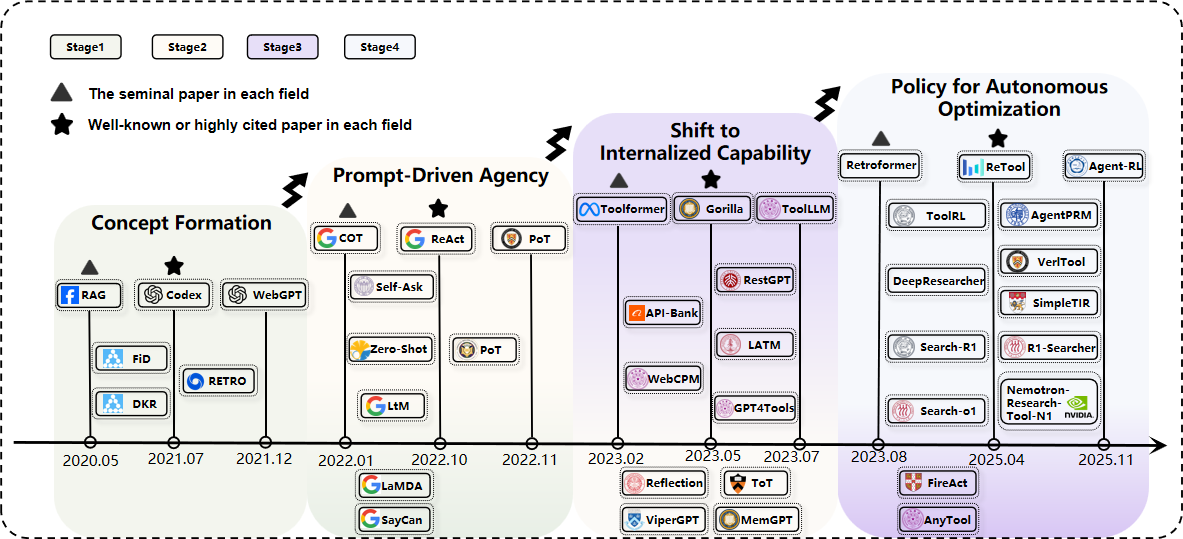}
  \caption{Timeline of Agentic Tool Use.}
  \Description{A timeline illustrating key milestones in the development of
  agent tool use, from early conceptual work to modern autonomous agents.}
  \label{fig:tool-use-timeline}
\end{figure}

In this section, we delineate the developmental trajectory of agentic tool use,
tracing the evolution from rudimentary neuro-symbolic concepts to the
sophisticated, self-optimizing agents of today. This progression encompasses
the initial theoretical blueprints, the emergence of prompt-based reasoning
frameworks, and the shift towards data-driven internalization and reinforcement
learning. The key milestones that define this timeline are illustrated in
Figure~\ref{fig:tool-use-timeline}.

\subsection{Early Exploration and Concept Formation}
The conceptual foundations of agentic tool use emerged before the rise of modern instruction-tuned LLMs. As researchers recognized that purely parametric models were limited in factual updating, precise computation, and interaction with external environments, early work began to augment language models with external resources rather than relying on scaling alone. One important step was Retrieval-Augmented Generation (RAG)~\cite{Lewis2020RAG}, which showed that coupling a generator with non-parametric memory could improve knowledge access and reduce hallucinations. This line of thought was further generalized by MRKL~\cite{Karpas2022MRKL}, which framed the language model as a central router that could dispatch subproblems to specialized modules and symbolic systems.

At the same time, several works moved beyond static augmentation toward explicit interaction. WebGPT~\cite{Nakano2021WebGPT} demonstrated that a model could browse the web to gather evidence for long-form question answering, while LaMDA~\cite{Thoppilan2022LaMDA} incorporated tools such as a calculator and a translator into dialogue generation, showing the value of grounding responses in external utilities. Beyond information access, TALM~\cite{Parisi2022TALM} explored how smaller models could be trained to generate tool calls and use returned results through supervised learning, providing an early instance of API-oriented tool use.

The notion of tools also expanded beyond conventional APIs. Codex~\cite{Chen2021Codex} suggested that language models could solve problems by generating executable code, effectively treating program execution as an external computational interface. In parallel, SayCan~\cite{Ahn2022DoAsICan} extended this paradigm to embodied settings by combining language-model priors with robotic affordances, allowing LLMs to guide physical action selection. These early studies did not yet constitute general-purpose agents, but they established the key ingredients of later systems: external knowledge access, modular routing, executable actions, and grounded interaction with digital or physical environments.

\subsection{The Emergence of Prompt-Driven Agency (2022-2023)}

Building on these early explorations, researchers began to observe that large language models could exhibit effective tool-use behavior without modifying their parameters. This stage was characterized by prompt engineering and in-context learning, which made it possible to elicit agent-like behavior from frozen models through suitable reasoning and interaction scaffolds. Chain-of-Thought prompting \cite{Wei2022ChainOfThought} laid the foundation by showing that LLMs could solve complex tasks through explicit intermediate reasoning. Building on this idea, ReAct \cite{Yao2023ReAct} connected reasoning with external action through a Thought-Action-Observation loop, allowing models to decide when to call tools and update subsequent steps based on returned observations. Self-Ask \cite{Press2023CompositionalityGap} further highlighted the value of decomposing complex questions into simpler subproblems, especially in tool-assisted question answering. As tasks became more challenging, Reflexion \cite{Shinn2023Reflexion} introduced self-reflective verbal feedback to help agents learn from past failures, while Tree of Thoughts \cite{Yao2023TreeOfThoughts} extended linear reasoning into search over multiple candidate trajectories, improving planning flexibility.

In parallel, another line of work improved reliability by shifting parts of reasoning to executable intermediates. Program-Aided Language Models \cite{Gao2023PAL} and Program of Thoughts \cite{Chen2023ProgramOfThoughts} used code generation to externalize computation, improving precision on arithmetic and symbolic problems. This idea was extended to multimodal and neural-symbolic settings by ViperGPT \cite{Suris2023ViperGPT} and Binder \cite{Cheng2023BindingSymbolic}, where code served as a structured interface for coordinating external tools. As tool ecosystems grew, researchers also explored more modular forms of orchestration. ART \cite{Paranjape2023ART} selected from tool-augmented reasoning templates, Chameleon \cite{Lu2023Chameleon} treated tools as composable modules, ReWOO \cite{Xu2023ReWOO} decoupled planning from execution to reduce interaction overhead, and HuggingGPT \cite{Shen2023HuggingGPT} framed the LLM as a central controller over diverse expert models.

Prompt-driven agency also began to move beyond single-turn task solving toward persistent and open-ended behavior. Generative Agents \cite{Park2023GenerativeAgents} demonstrated the role of memory in sustained agent behavior, MemGPT \cite{Packer2023MemGPT} introduced a memory management mechanism for handling long-context interaction, and Voyager \cite{Wang2023Voyager} showed that prompt-based agents could continuously acquire and refine skills in embodied environments. These works marked a distinct phase in which tool use, planning, and memory could be elicited primarily through prompting, establishing the practical foundation for later training-based and policy-driven agent systems.

\subsection{The Shift to Supervised Tool Learning (2023-2024)}

As prompt-based tool use became increasingly capable, its limitations also became more apparent. Repeated prompting introduced substantial latency and token overhead, while long reasoning-action trajectories often remained brittle in complex tasks. These challenges motivated a transition from eliciting tool use in context to internalizing it through training, allowing models to invoke tools more efficiently and reliably. Early efforts explored supervised imitation of tool-augmented behavior. WebCPM \cite{Qin2023WebCPM} showed that language models could be fine-tuned on human web-browsing trajectories for goal-directed interaction, and Toolformer \cite{Schick2023Toolformer} further demonstrated that models could teach themselves when and how to call APIs by inserting tool-use annotations into their own training data. This direction was extended to instruction-tuned settings by ToolAlpaca \cite{Tang2023ToolAlpaca}, which used synthetic supervision from stronger models, and by ToolLLM \cite{Qin2023ToolLLM}, which scaled training to thousands of real-world APIs.

As the number and diversity of tools increased, internalized tool use also required better mechanisms for selection, adaptation, and evaluation. API-Bank \cite{Li2023APIBank} provided an early benchmark for planning, tool calling, and response generation in API-based interaction. To improve scalability under large tool libraries, AnyTool \cite{Du2024AnyTool} introduced hierarchical retrieval to narrow the candidate space before invocation. The dynamic nature of external interfaces posed another challenge, since APIs evolve over time and new tools appear continuously. Gorilla \cite{Patil2024Gorilla} addressed this issue by combining retrieval with API-aware fine-tuning, enabling models to use up-to-date documentation at inference time. Related efforts such as RestGPT \cite{Song2023RESTGPT} and GPT4Tools \cite{Yang2023GPT4Tools} further explored structured interaction with RESTful APIs and the generation of specialized tool-use data.

Later work increasingly focused on robustness, transfer, and the balance between agentic behavior and general capability. FireAct \cite{Chen2023FireAct} showed that fine-tuning on interleaved reasoning-action trajectories could substantially improve the performance of smaller open models. Generalization to unseen tools was further studied by NexusRaven \cite{Srinivasan2023NexusRaven}, which emphasized accurate function calling from tool descriptions, while APIGen \cite{Liu2024APIGen} improved data quality through verifiable synthetic API-call generation. At the same time, researchers began to examine how tool specialization could be integrated with broader model competence. Lemur \cite{Xu2024Lemur} proposed a unified framework that balanced function calling, tool planning, and general language understanding, and AgentTuning \cite{Zeng2024AgentTuning} further highlighted the importance of preserving generalist reasoning ability while optimizing for agentic behavior. These studies marked the transition from learning tool use through prompts to learning it in model weights, establishing supervised fine-tuning and synthetic data generation as practical routes toward more efficient, scalable, and robust tool-using agents.

\subsection{The Rise of Policy for Autonomous Optimization (2024-Present)}

Despite the effectiveness of supervised fine-tuning, an important limitation remained: static training signals were often insufficient for exploration, error recovery, and long-horizon adaptation in dynamic environments. As a result, recent research has increasingly turned to reinforcement learning, treating multi-turn tool use as a sequential decision-making problem optimized through environmental feedback rather than imitation alone. This stage can be understood as a shift from learning how to call tools to learning how to optimize interaction with them.

Early efforts in this direction focused on improving the quality of tool-use decisions under feedback. ReTool \cite{Feng2025ReTool} applied reinforcement learning to help agents distinguish between cases that required external tools and those that could be solved from internal knowledge, thereby reducing unnecessary tool calls. Another important challenge was how to learn from failed executions. Retroformer \cite{Yao2024Retroformer} addressed this issue through retrospective learning over past tool-use trajectories, enabling agents to revise policies based on environmental outcomes. Along a similar line, TRICE \cite{Qiao2024TRICE} tackled credit assignment in long tool chains by strengthening steps that contributed to correct final answers while suppressing noisy or unhelpful actions.

More recent work has moved toward jointly optimizing reasoning and tool-use policies at scale. Search-R1 \cite{Jin2025SearchR1} showed that reinforcement learning over the interaction between reasoning and retrieval could induce more effective search behavior than imitation-based methods alone. SimpleTIR \cite{Xue2025SimpleTIR} further unified reasoning and tool execution within a single end-to-end training loop, reducing the mismatch often introduced by pipeline-style designs. At the same time, RLFactory \cite{Chai2025RLFactory} and VerlTool \cite{Jiang2025VerlTool} helped systematize this emerging paradigm by providing infrastructure for trajectory-level reward modeling and broader agentic reinforcement learning workflows. These studies suggest an emerging stage in which tool-using agents are optimized not only to invoke external resources, but also to refine their interaction strategies through feedback and trial-and-error.

\section{Prompting as plug-and-play}
\label{sec:prompting-plug-and-play}

\begin{figure}[!htbp]
  \centering
  \includegraphics[width=\linewidth]{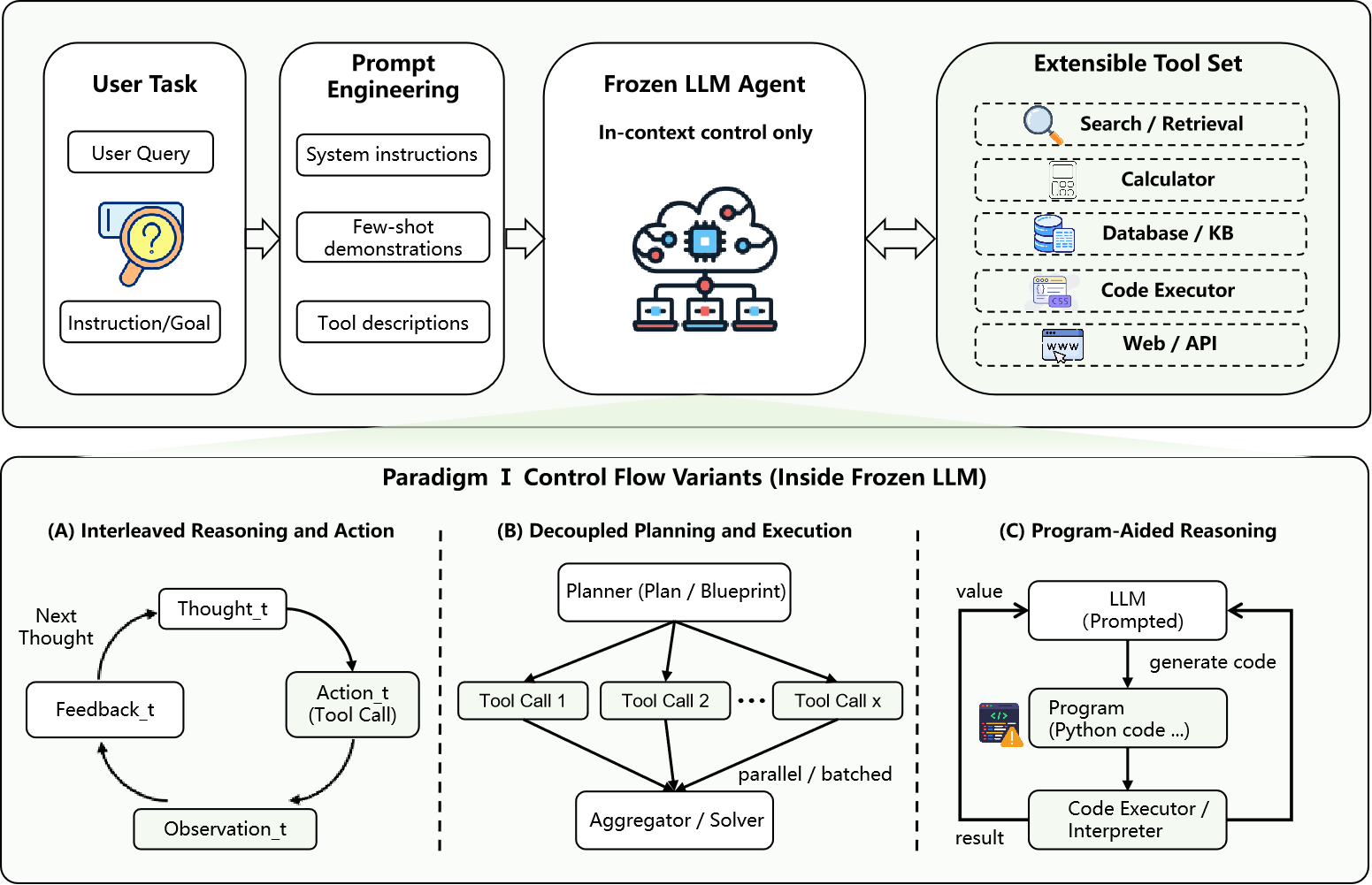}
  \caption{Illustration of the Prompting as plug-and-play paradigm.}
  \label{fig:prompting-as-plug-and-play}
\end{figure}

Prompting as plug-and-play refers to tool-use methods that rely on prompting and in-context interaction rather than parameter updates. In this paradigm, the language model remains frozen and serves as the central reasoning and coordination module, while tool-use behavior is elicited through instructions, demonstrations, and feedback from external environments. This design offers several practical advantages, including flexibility, interpretability, and low deployment cost, making it particularly suitable for rapid prototyping and adaptation to newly introduced tools. At the same time, because the model does not internalize tool-use behavior in its parameters, these methods often suffer from higher latency and less stable performance in complex or long-horizon tasks. To better organize this paradigm, we group existing methods into three representative structures according to how reasoning, planning, and execution are coordinated: Interleaved Reasoning and Action, Decoupled Planning and Execution, and Program-Aided Reasoning. As illustrated in Figure~\ref{fig:prompting-as-plug-and-play}, these three categories capture the main control flows through which frozen LLMs interact with external tools.

\subsection{Interleaved Reasoning and Action}
\label{subsec:interleaved-reasoning-action}

A representative structure within Paradigm I is interleaved reasoning and action, which treats problem solving as a multi-turn interaction between the agent and its environment. Rather than completing all reasoning before acting, the model alternates between intermediate deliberation and external operations, using observations from the environment to guide subsequent decisions. A representative starting point in this line is ReAct~\cite{Yao2023ReAct}, which interleaves reasoning traces with task-specific actions so that the model can explicitly decide what to do next based on newly acquired information. This design helps mitigate the disconnect between abstract reasoning and external execution, and has become a canonical control loop for prompting-based tool use.

Subsequent work extended this interleaved loop in several directions. One line focused on improving robustness through self-correction and iterative refinement. Reflexion~\cite{Shinn2023Reflexion} augments the standard interaction loop with verbal self-reflection, allowing the agent to critique failed trajectories and reuse these reflections in later attempts, while Self-Refine~\cite{Madaan2023SelfRefine} shows that a model can iteratively improve its own intermediate reasoning or final outputs through self-generated feedback. Another line introduced explicit verification mechanisms. Chain-of-Verification~\cite{Dhuliawala2024ChainOfVerification} prompts the model to generate verification questions for its own claims before producing a revised answer, and CRITIC~\cite{Gou2024Critic} further grounds this process in external tools by validating intermediate or final outputs with search engines and other utilities. To better handle tasks that require foresight and exploration, LATS~\cite{Zhou2023LATS} combines the interaction loop of ReAct with tree search, enabling more deliberate exploration of alternative reasoning-action trajectories.

Related prompting techniques can further strengthen this paradigm, although they are not themselves interleaved tool-use frameworks. Self-Consistency~\cite{Wang2023SelfConsistency} improves robustness by aggregating multiple sampled reasoning paths, Least-to-Most Prompting~\cite{Zhou2023LeastToMost} encourages decomposition into simpler subproblems, and Take a Step Back~\cite{Zheng2023TakeAStepBack} promotes abstraction before low-level action selection. These methods show how prompting alone can support a tightly coupled loop of reasoning, acting, observing, and revising, making interleaved control one of the most influential architectures in prompting-based agentic tool use.

\subsection{Decoupled Planning and Execution}
\label{subsec:decoupled-planning-execution}

A second representative structure within Paradigm I is decoupled planning and execution, which separates high-level reasoning from downstream tool invocation. Compared with interleaved reasoning-action loops, this design reduces repeated deliberation at every step and can improve efficiency when tasks require long tool chains or multiple dependent operations. Instead of alternating between thought and action after each observation, the model first produces a structured plan and then executes it through external tools, allowing planning and execution to be handled more explicitly.

A representative starting point in this line is ReWOO~\cite{Xu2023ReWOO}, which decouples reasoning from observation by introducing a planner-worker architecture. The planner first generates a sequence of executable subtasks with variable assignments, and the worker then carries them out without requiring the model to replan after every intermediate observation. This design reduces interaction overhead and makes tool use more efficient. Subsequent work further extended this idea toward richer scheduling and more adaptive control. LLMCompiler~\cite{Kim2024LLMCompiler} organizes tool calls as a dependency-aware computation graph and enables parallel execution when subtasks are independent, while AdaPlanner~\cite{Sun2023AdaPlanner} shows that decoupled planning can still remain adaptive by revising plans in response to environmental feedback.

Related methods explored neighboring forms of plan-first control. Plan-and-Solve Prompting~\cite{Wang2023PlanAndSolve} highlights the value of generating an explicit plan before solving a task, and Skeleton-of-Thought~\cite{Ning2023SkeletonofThought} similarly improves efficiency by producing a coarse outline before expanding details. In more structured settings, PEARL~\cite{Sun2024PEARL} applies a plan-and-execute strategy to long documents, allowing the model to decompose document-grounded tasks into manageable actions. Other systems extended decoupled control to broader tool orchestration. ART~\cite{Paranjape2023ART} retrieves suitable reasoning-action programs from an external task library, Chameleon~\cite{Lu2023Chameleon} composes tools through modular execution pipelines, and systems such as HuggingGPT~\cite{Shen2023HuggingGPT} and TaskMatrix.AI~\cite{Liang2024TaskMatrixAI} further illustrate how an LLM can act as a central planner over diverse expert models and APIs. These methods show that separating planning from execution offers a practical alternative to tightly interleaved control, especially when tool use requires structured decomposition, reduced latency, or coordination across multiple modules.

\subsection{Program-Aided Reasoning}
\label{subsec:program-aided-reasoning}

The third sub-paradigm addresses the tendency of LLMs to make arithmetic and logical errors when reasoning purely in natural language by shifting the intermediate representation from text to executable code. In this setting, the external program executor is not merely an auxiliary utility, but becomes part of the reasoning process itself. PAL~\cite{Gao2023PAL} and PoT~\cite{Chen2023ProgramOfThoughts} are representative starting points in this line, prompting the model to generate Python programs that externalize computation and decouple reasoning from execution. To further strengthen logical rigor, Logic-LM~\cite{Pan2023LogicLM} translates natural language problems into symbolic formulations that can be solved by external logic solvers, improving determinism in deductive reasoning tasks.

This program-centric perspective was later extended to broader execution environments. Chain of Code~\cite{Li2024ChainOfCode} broadens code-based reasoning beyond purely executable computation by allowing the model to mix runnable code with flexible pseudo-code, so that semantic sub-tasks can still be handled through a language-model-augmented code emulator. ViperGPT~\cite{Suris2023ViperGPT} applies a similar philosophy to visual reasoning, where the model generates Python code to call vision APIs and compose multimodal operations. In embodied settings, Code as Policies~\cite{Liang2022CodeAsPolicies} further shows that natural language instructions can be compiled into robot policy code through predefined control APIs, extending program-aided reasoning into action generation.

The same idea also motivates more advanced forms of tool creation and reuse. MathPrompter~\cite{Imani2023MathPrompter} improves reliability by generating multiple candidate Python functions and checking for agreement across their execution results. LATM~\cite{Cai2024LLMToolMakers} goes one step further by enabling agents to create reusable Python tools for unseen tasks, while CREATOR~\cite{Qian2023CREATOR} further formalizes this process by separating tool creation from downstream decision execution. These methods show that prompting can use code not only as an execution backend, but also as a structured medium for reasoning, control, and even tool invention itself.

\section{Supervised Tool Learning}
\label{sec:training-internalized}

\begin{figure}[!htbp]
  \centering
  \includegraphics[width=\linewidth]{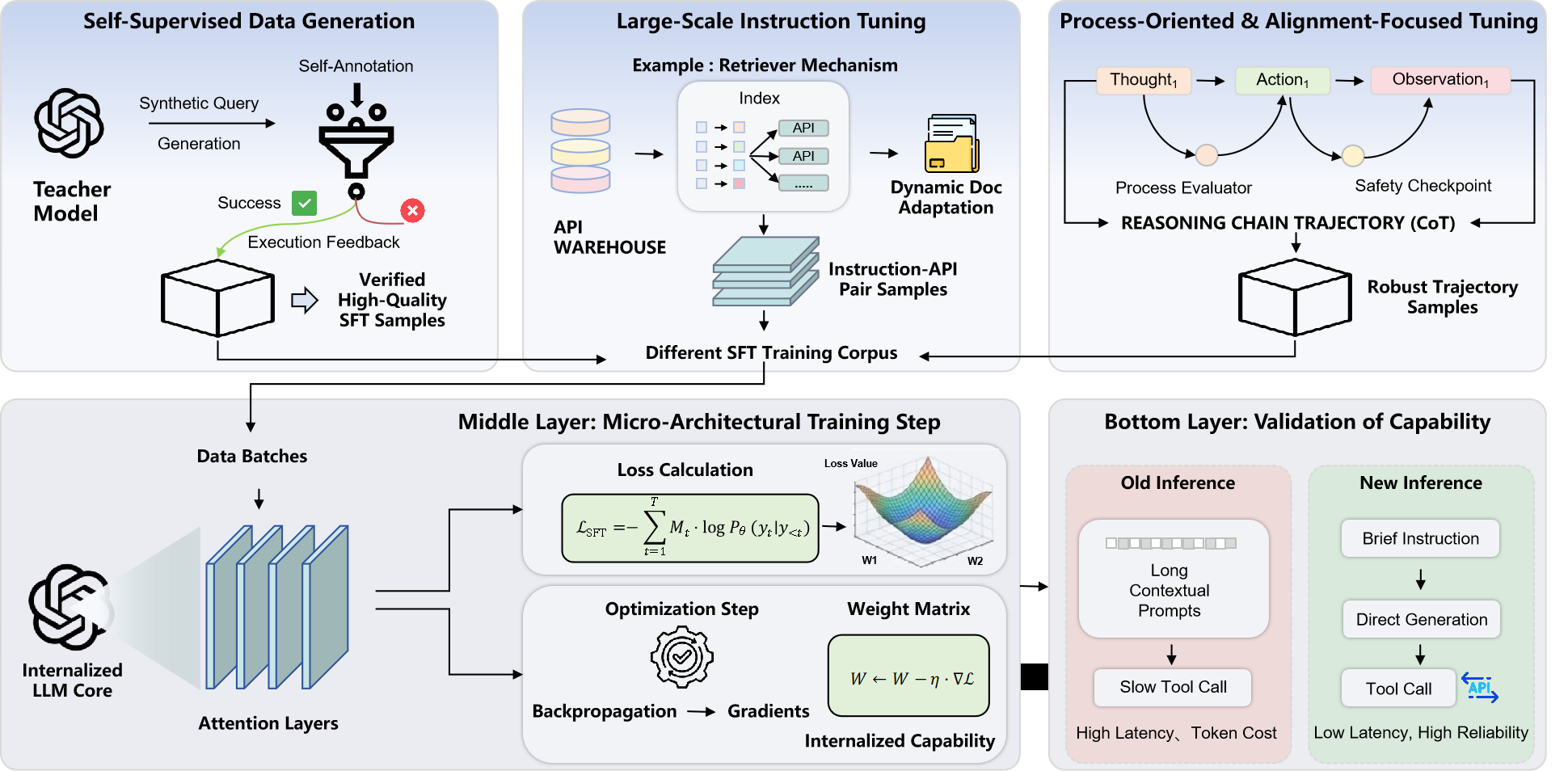}
  \caption{Training Pipeline of Paradigm II: From Supervision to Internalized Tool-Use Capability.}
  \Description{A schematic illustration of supervised fine-tuning workflows that internalize
  tool-use capabilities into model parameters.}
  \label{fig:training-internalized}
\end{figure}

Compared with prompting-only approaches, this paradigm shifts the focus from externally eliciting tool-use behavior at inference time to internalizing it through training. Rather than relying on prompts alone to scaffold reasoning and tool invocation, Supervised Tool Learning uses labeled or synthetic supervision to encode tool-use patterns directly into model parameters. This transition improves efficiency, stability, and scalability, while also enabling models to generalize more reliably across diverse tools and task settings. In this setting, the central challenge is not only how to obtain training data at scale, but also how to design supervision signals that capture tool syntax, invocation semantics, planning structure, and alignment requirements. As illustrated in Figure~\ref{fig:training-internalized}, these categories together show how different forms of supervision enable models to acquire and internalize tool-use capability.

\subsection{Self-Supervised Data Generation}
\label{subsec:Self-Supervised Data Generation}

A natural starting point for Paradigm II is to reduce the dependence on costly human annotation by automatically constructing tool-use supervision. Early efforts explored self-supervised data generation, where models learn tool-use patterns by labeling or simulating their own interaction traces rather than relying entirely on manually curated demonstrations. A representative starting point in this line is Toolformer~\cite{Schick2023Toolformer}, which showed that a language model could annotate its own training corpus with API calls and corresponding outputs, thus learning when and how external tools should be invoked with minimal human supervision. More concretely, the key idea is to retain a candidate tool call only when its returned result improves next-token prediction under the base language modeling objective. This can be written as the reduction in prediction loss after inserting the tool response:
\begin{equation}
\Delta \ell_i(c,r)
=
\ell_i^{-}-\ell_i^{+}
=
-\log p_{\theta}(x_i \mid c)
+
\log p_{\theta}(x_i \mid c, r),
\label{eq:toolformer_gain}
\end{equation}
where $c$ denotes the original context, $r$ denotes the tool response, and $x_i$ denotes a subsequent token to be predicted. This criterion captures the central intuition of self-supervised tool annotation: a tool call is useful not merely because it is syntactically valid, but because it measurably improves the model's prediction of what comes next.

Subsequent work extended this idea from local self-annotation to richer synthetic trajectory generation. ToolAlpaca~\cite{Tang2023ToolAlpaca} used synthetic tool-use demonstrations generated by stronger models to train smaller open-source LLMs, showing that instruction-tuned tool-use behavior could be effectively transferred through generated corpora. AutoAct~\cite{Qiao2023AutoAct} further reduced reliance on handcrafted demonstrations by automatically synthesizing planning trajectories for tool-augmented agents, making it possible to bootstrap agentic behavior from limited initial supervision.

As the focus shifted from trajectory generation to function-calling quality, later work increasingly emphasized verifiability and data quality control. APIGen~\cite{Liu2024APIGen} introduced a multi-stage pipeline for generating and validating API-call training data, improving the reliability of synthetic supervision through executable verification. ToolACE~\cite{Liu2024ToolACE} further developed this direction with a self-evolving framework that combines API augmentation, dialogue synthesis, and hierarchical validation to produce higher-quality tool-use data. Confucius~\cite{Gao2024Confucius} complements these efforts by incorporating introspective feedback and curriculum-style refinement, highlighting that synthetic data generation can be improved not only by scale, but also by progressively structured supervision. These methods show that self-supervised and synthetic data generation provides a practical foundation for training tool-using models at scale, especially when manual collection of high-quality trajectories is expensive or infeasible.

\subsection{Large-Scale Instruction Tuning}
\label{subsec:large-scale-instruction-tuning}

Once synthetic and self-supervised supervision became feasible, the focus of Paradigm II shifted from data construction to scale and generalization. The goal was no longer to train agents for a small set of predefined tools, but to build models that could handle thousands of diverse real-world APIs with stronger generalization across domains and interfaces. A representative turning point in this direction is ToolLLM~\cite{Qin2023ToolLLM}, which introduced ToolBench, a large-scale dataset covering more than 16,000 real-world RESTful APIs, and showed that instruction tuning on broad API corpora could substantially expand the scope of tool-use capability.

As tool libraries became larger and more dynamic, large-scale tuning also required better support for retrieval, adaptation, and tool selection. Gorilla~\cite{Patil2024Gorilla} addressed the problem of API evolution through retrieval-aware training, enabling models to condition tool use on retrieved documentation rather than relying solely on static parametric memory. API-Bank~\cite{Li2023APIBank} complemented this line by emphasizing the planning and retrieval abilities needed to select appropriate tools from large candidate pools. To further improve scalability, AnyTool~\cite{Du2024AnyTool} introduced hierarchical retrieval over large toolsets, while EasyTool~\cite{Yuan2025EasyTool} reduced context overhead by transforming lengthy documentation into more token-efficient representations.

Recent work has increasingly focused on improving invocation efficiency and generalization in practical function-calling settings. Octopus v2~\cite{Chen2024OctopusV2} optimized latency through a lightweight mapping strategy for function arguments, and ToolkenGPT~\cite{Hao2023ToolkenGPT} represented tools as parametric token embeddings to support large tool collections without excessively expanding the vocabulary. For more specialized high-performance function calling, xLAM~\cite{Zhang2025xLAM} targeted large action spaces, while Granite-Function-Calling~\cite{Abdelaziz2024GraniteFunctionCalling} improved parameter prediction through multi-task learning. Functionary~\cite{MeetKai2024Functionary} further emphasized the tight integration of conversational interaction and function invocation, and NexusRaven~\cite{Srinivasan2023NexusRaven} highlighted robust zero-shot generalization to unseen tool definitions. These methods show how large-scale instruction tuning moves beyond isolated tool-use demonstrations toward general-purpose models that can retrieve, interpret, and invoke broad tool ecosystems with increasing efficiency and reliability.

\subsection{Process-Oriented and Alignment-Focused Tuning}
\label{subsec:process-oriented-alignment}

The third sub-paradigm moves beyond the simple correctness of function calls to optimize the reasoning process that leads to tool use and to align that process with human preferences and safety requirements. Executing a tool is not sufficient if the underlying rationale is flawed, brittle, or misaligned with the intended objective. A representative early effort in this direction is FireAct~\cite{Chen2023FireAct}, which showed that fine-tuning on reasoning trajectories can better internalize problem-solving structure than direct input-output supervision alone. ToRA~\cite{Gou2024ToRA} further strengthens this line by training on interleaved reasoning-action trajectories with explicit rationales before each tool call, helping bridge the semantic gap between language reasoning and executable actions. Synapse~\cite{Zheng2024Synapse} extends this idea by distilling agent trajectories into exemplars that can guide subsequent decision making.

Another line of work focuses on making these reasoning processes more robust and discriminative. Masked Thought~\cite{Chen2024MaskedThought} introduces a training objective that masks parts of the reasoning chain and asks the model to recover the missing logic, thereby discouraging superficial pattern matching. Agent-FLAN~\cite{Chen2024AgentFLAN} decomposes agent capability into format following and reasoning, and uses negative samples to reduce hallucinated actions. In retrieval-dependent settings, RAFT~\cite{Zhang2024RAFT} further improves robustness by training models to distinguish relevant from irrelevant retrieved documents, which is particularly useful when tool use depends on external information sources.

Alignment and safety then become increasingly important once tool-using models are deployed in realistic environments. ToolAlign~\cite{Chen2024ToolAlign} introduces the H2A principle, balancing helpfulness, harmlessness, and autonomy in tool-use supervision. ToolSword~\cite{Ye2024ToolSword} targets safety alignment more directly by constructing adversarial scenarios such as prompt injection attacks for robust tool-use training, while RoT~\cite{Ye2024RoTBench} improves stability under noisy conditions by exposing models to imperfect tool documentation during training. MetaTool~\cite{Huang2024MetaTool} complements these efforts by training models to assess whether a tool is actually needed before attempting invocation. Finally, several works emphasize that tool specialization should not come at the expense of broader competence. Lemur~\cite{Xu2024Lemur} balances function calling, planning, and general language understanding within a unified framework, and AgentTuning~\cite{Zeng2024AgentTuning} preserves generalist ability by mixing agent-oriented supervision with ordinary conversational data. These methods show that tuning for tool use must optimize not only whether a model can call tools, but also how it reasons, when it acts, and whether its behavior remains aligned, robust, and broadly capable.

\section{Reward-Driven Tool Policy Learning}
\label{sec:policy-autonomous}

\begin{figure}[!htbp]
    \centering
    \includegraphics[width=\linewidth]{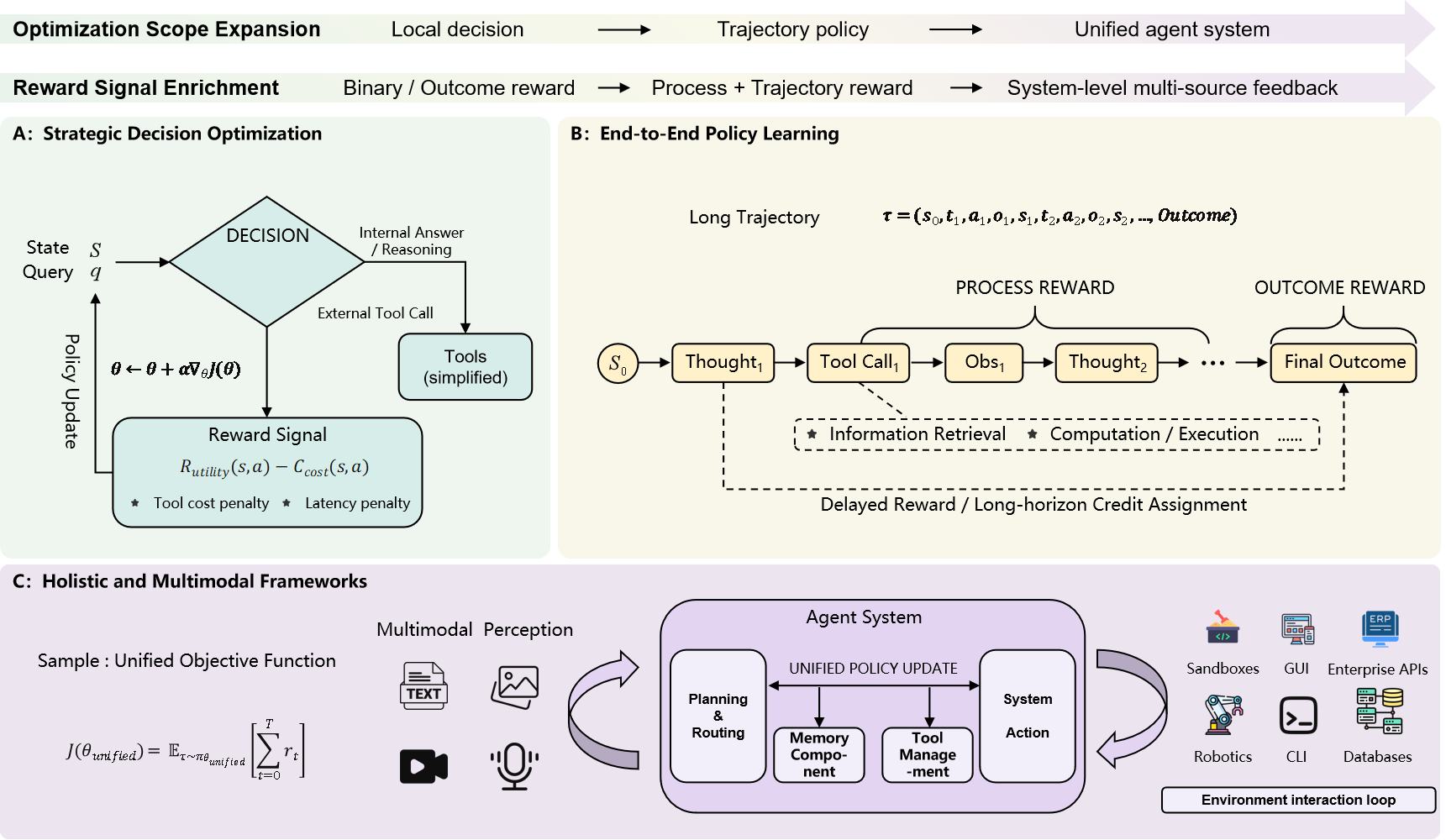}
    \caption{An overview of Paradigm~III: Reward-Driven Tool Policy Learning.}
    \label{fig:policy-autonomous}
\end{figure}

Compared with the previous paradigm, which mainly internalizes tool-use behavior through static supervision, this paradigm focuses on optimizing tool interaction through environmental feedback. The central challenge is no longer only whether a model can call a tool correctly, but also when it should act, how it should recover from failures, and how it can improve its strategy over multi-turn interactions. For this reason, recent research has increasingly turned to reinforcement learning, treating tool use as a sequential decision-making problem rather than a purely imitative one. To organize this paradigm, we group existing methods into three categories according to the dominant scope of policy optimization: Strategic Decision Optimization, End-to-End Policy Learning for Multi-Turn Reasoning, and Holistic and Multimodal Agentic Frameworks. As illustrated in Figure~\ref{fig:policy-autonomous}, these categories reflect a progression from optimizing individual tool-use decisions to training more general policies for long-horizon and multimodal agent behavior.

\subsection{Strategic Decision Optimization}
\label{subsec:Strategic Decision Optimization}
The earliest application of reinforcement learning in this paradigm focused on high-level strategic decisions about tool use. Rather than optimizing the entire reasoning-action trajectory, this line of work asks a more basic but important question: when should an agent invoke an external tool, and when should it rely on its internal knowledge instead? The objective is not only to improve accuracy, but also to balance tool effectiveness against computational cost and latency. In practice, such strategic tool-use policies are often optimized with policy-gradient style reinforcement learning objectives. A widely used method is the PPO clipped surrogate objective:
\begin{equation}
\mathcal{L}^{\mathrm{CLIP}}(\theta)
=
\hat{\mathbb{E}}_t
\left[
\min\!\left(
r_t(\theta)\hat{A}_t,\,
\mathrm{clip}\!\left(r_t(\theta),1-\epsilon,1+\epsilon\right)\hat{A}_t
\right)
\right],
\label{eq:ppo_clip}
\end{equation}
where
\begin{equation}
r_t(\theta)=\frac{\pi_{\theta}(a_t\mid s_t)}{\pi_{\theta_{\mathrm{old}}}(a_t\mid s_t)}.
\label{eq:ppo_ratio}
\end{equation}
Here, $\hat{A}_t$ denotes the estimated advantage of a tool-use action, and the clipping operation stabilizes policy updates by preventing overly large deviations from the previous policy. In the context of strategic decision optimization, it captures how an agent can gradually learn whether invoking a tool at a given state is beneficial under reward feedback.

A representative starting point in this direction is ReTool~\cite{Feng2025ReTool}, which formulates tool invocation as a reinforcement learning problem and focuses on learning when external tools should be used during reasoning. Rather than assuming that more tool use is always beneficial, it optimizes a reward signal that favors correct outcomes while discouraging unnecessary calls, thereby improving the selectivity and efficiency of tool-assisted reasoning. ToolRL~\cite{Qian2025ToolRL} further simplifies this setting by showing that carefully scaled outcome-based rewards can already provide an effective signal for optimizing tool-use behavior, reducing the need for more elaborate reward engineering.

Subsequent work extended strategic optimization toward capability expansion and self-improvement. ToolExpander~\cite{Chen2025ToolExpander} introduces a dynamic multi-round hard-sampling strategy that identifies persistent failure cases and replaces them with stronger demonstrations during reinforcement learning, effectively serving as a curriculum for weaker models. Agent0~\cite{Xia2025Agent0} moves further in this direction by allowing the agent to act as both curriculum designer and executor, using Ambiguity Dynamic Policy Optimization (ADPO) to generate tasks near the frontier of its current ability and improve without external data. In more specialized settings, Arcee~\cite{Wang2025LargeActionModels} extends this idea to domain-specific workflows through a Large Action Model trained with reinforcement learning, showing that strategic policy optimization can also support structured enterprise action spaces. These methods focus less on full end-to-end autonomy than on improving high-level decisions about whether, when, and how agents should act.

\subsection{End-to-End Policy Learning for Multi-Turn Reasoning}
\label{End-to-End Policy Learning for Multi-Turn Reasoning}
Compared with strategic decision optimization, this line of work directly optimizes the full reasoning-action policy over multi-turn interactions. The key challenge is no longer only whether an agent should invoke a tool, but how to assign credit across long trajectories where an early mistake may propagate through later reasoning and execution steps. SearchR1 \cite{Jin2025SearchR1} is a representative example in this direction, showing that large-scale reinforcement learning can train agents to develop search-oriented reasoning strategies such as query reformulation and information synthesis from sparse outcome rewards alone. SimpleTIR \cite{Xue2025SimpleTIR} further pushes this setting toward end-to-end optimization by jointly training reasoning and tool execution within a unified reinforcement learning loop, reducing the mismatch introduced by pipeline-style designs. Earlier efforts such as Retroformer \cite{Yao2024Retroformer} and TRICE \cite{Qiao2024TRICE} provide important foundations for this line by enabling agents to learn from retrospective execution feedback and by improving credit assignment in long tool-use chains.

A second line of research seeks to make this optimization more stable by introducing denser or more structured learning signals. AgentPRM~\cite{Xi2025AgentPRM} adapts process reward modeling to agentic tasks, evaluating intermediate steps in terms of both their promise and their progress toward the goal. LeTS~\cite{Zhang2025LeTS} similarly combines process-level and outcome-level rewards to guide the joint development of reasoning and retrieval behavior. RAGEN~\cite{Wang2025RAGEN} extends this perspective to self-evolving agents by generating reflective revision data from prior trajectories and using it to reinforce error-correcting behaviors during multi-turn reinforcement learning.

More recent work explores richer coordination mechanisms within end-to-end policy learning. MATPO~\cite{Mo2025MATPO} introduces role-aware optimization for planner-worker coordination, distributing rewards across distinct functions within the same agent framework so that delegation and execution can be improved simultaneously. Agent-R~\cite{Yuan2025AgentR} further treats reflection as an explicit action inside the reinforcement learning loop, allowing the agent to critique past behavior and incorporate those critiques into future policy updates. In this setting, reinforcement learning is used not merely to refine isolated tool decisions, but to shape how agents search, recover, coordinate, and revise their behavior over extended interaction horizons.

\subsection{Holistic and Multimodal Agentic Frameworks}
\label{subsec:holistic-multimodal}

The frontier of Paradigm III is expanding beyond text-based APIs to encompass multimodal interactions and holistic system optimization. As agents are deployed in rich visual environments, the definition of tool use evolves to include visual perception, GUI manipulation, and long-term memory management.

VerlTool~\cite{Jiang2025VerlTool} proposes a holistic framework for agentic reinforcement learning, arguing that optimizing tool calls in isolation is insufficient. It treats the entire agent system, including its memory module and tool selector, as a unified policy and updates all components jointly. DeepAgent~\cite{Li2025DeepAgent} complements this perspective by introducing an autonomous memory folding mechanism that compresses past interactions into structured episodic memory to maintain coherence over long horizons. AgentRL~\cite{Zhang2025AgentRL} extends this line toward scalable multi-turn and multi-task training through an asynchronous generation-training pipeline, a unified function-call based API interface, and a centralized environment controller, highlighting the growing importance of system-level infrastructure in agentic reinforcement learning. RLFactory~\cite{Chai2025RLFactory} further supports this trend by providing a plug-and-play infrastructure for simulating tool environments and computing trajectory-level rewards, facilitating the development of self-evolving agents.

In the domain of visual and embodied agents, Agent Q~\cite{Putta2024AgentQ} combines guided Monte Carlo Tree Search (MCTS) with a self-critique mechanism, allowing agents to learn from both successful and unsuccessful trajectories in interactive web environments through self-play. SEAgent~\cite{Sun2025SEAgent} focuses on computer-use agents that learn from experience, employing a self-evolution mechanism in which the agent autonomously explores the OS environment and curates its own instruction-tuning data. DigiRL~\cite{Bai2024DigiRL} addresses the challenge of training in-the-wild device control agents by introducing a scalable offline-to-online reinforcement learning framework that allows agents to refine their GUI interaction policies through autonomous exploration on Android devices.

Advanced frameworks also incorporate structural priors and multimodal grounding. GEPO~\cite{Yuan2025GraphEnhancedPolicyOpt} dynamically constructs a state-transition graph from agent experience and uses graph centrality to guide exploration toward high-impact states in open-ended environments. MGPO~\cite{Huang2025MGPO} specifically targets visual reasoning, enabling agents to refine their focus on key image regions through sequential decision-making. OS-Copilot~\cite{Wu2024OSCopilot} presents a framework for building generalist computer agents that optimize user intent satisfaction across diverse applications. Cradle~\cite{Tan2025Cradle} extends this idea to general computer control, enabling agents to operate software through screen observation and keyboard-mouse emulation. ToolRM~\cite{Agarwal2025ToolRM} complements these frameworks by introducing specialized outcome reward models for evaluating the efficiency and correctness of tool-use trajectories, providing stronger learning signals for complex agentic reinforcement learning.

\section{Evaluating Agentic Tool Use: Benchmarks and Metrics}
\label{sec:evaluation}

The evaluation of agentic tool use has evolved alongside the underlying methodological paradigms. Early benchmarks focused on whether models could generate valid and schema-compliant tool calls, capturing the correctness of tool invocation at the function level. As tool use became more tightly integrated with reasoning and planning, evaluation expanded to end-to-end task success on static datasets. More recent benchmarks further move toward interactive and high-fidelity environments, where agents must adapt to changing states, recover from execution errors, and satisfy safety constraints. Following this progression, we organize existing evaluation benchmarks into three levels: tool usage correctness, task completion, and tool-driven interaction.

\subsection{Tool Usage Correctness}
\label{subsec:tool-usage-correctness}

This category forms the diagnostic foundation of agentic tool-use evaluation. It
asks whether a model can understand tool specifications, determine whether tool
use is necessary, select the appropriate function from a candidate pool, and
produce structurally valid calls with correctly grounded arguments. In contrast
to the later layers of evaluation, benchmarks at this level primarily assess
local competence in tool invocation rather than end-to-end task success.
Representative benchmarks in this layer are summarized in
Table~\ref{tab:eval_tool_use_correctness}.

\begin{table}[!htbp]
  \centering
  \footnotesize
  \setlength{\tabcolsep}{2.0pt}
  \renewcommand{\arraystretch}{1.05}
  \caption{Representative benchmarks for tool usage correctness.}
  \label{tab:eval_tool_use_correctness}
  \rowcolors{2}{gray!10}{white}

  \begin{tabularx}{\columnwidth}{@{}
    >{\raggedright\arraybackslash}p{0.08\columnwidth}
    >{\raggedright\arraybackslash}p{0.18\columnwidth}
    >{\raggedright\arraybackslash}p{0.27\columnwidth}
    >{\centering\arraybackslash}p{0.12\columnwidth}
    >{\centering\arraybackslash}p{0.13\columnwidth}
    >{\raggedright\arraybackslash}p{0.15\columnwidth}
  @{}}
    \toprule
    \rowcolor{gray!15}
    \textbf{Year} &
    \textbf{Benchmark} &
    \textbf{Primary Focus} &
    \textbf{Executable} &
    \textbf{Compositional} &
    \textbf{Tool Source} \\
    \midrule

    \rowcolor{white}
    \multicolumn{6}{@{}l@{}}{\textbf{Category I: Function-Call Validity}} \\
    \addlinespace[2pt]

    2025 & \textbf{BFCL}~\cite{Patil2025BFCL} &
    Function-call validity &
    \xmark &
    \xmark &
    Curated functions \\

    2025 & \textbf{ToolEyes}~\cite{Ye2025ToolEyes} &
    Fine-grained error diagnosis &
    \xmark &
    \xmark &
    Curated functions \\

    2024 & \textbf{T-Eval}~\cite{Chen2024TEval} &
    Step-wise invocation correctness &
    \xmark &
    \cmark &
    Curated functions \\

    \midrule

    \rowcolor{white}
    \multicolumn{6}{@{}l@{}}{\textbf{Category II: Tool Retrieval and Selection}} \\
    \addlinespace[2pt]

    2023 & \textbf{API-Bank}~\cite{Li2023APIBank} &
    API retrieval and selection &
    \cmark &
    \cmark &
    Simulated tools \\

    2024 & \textbf{StableToolBench}~\cite{Guo2024StableToolBench} &
    Large-scale tool selection &
    \cmark &
    \cmark &
    Real-world APIs \\

    2024 & \textbf{Gorilla}~\cite{Patil2024Gorilla} &
    Retrieval over evolving APIs &
    \cmark &
    \xmark &
    Evolving API docs \\

    \midrule

    \rowcolor{white}
    \multicolumn{6}{@{}l@{}}{\textbf{Category III: Tool-Use Decisions}} \\
    \addlinespace[2pt]

    2025 & \textbf{When2Call}~\cite{Ross2025When2Call} &
    Tool-use necessity &
    \xmark &
    \xmark &
    Curated functions \\

    2024 & \textbf{WTU-Eval}~\cite{Ning2024WTUEval} &
    Whether-or-not tool usage &
    \xmark &
    \xmark &
    Curated functions \\

    \midrule

    \rowcolor{white}
    \multicolumn{6}{@{}l@{}}{\textbf{Category IV: Compositional Invocation}} \\
    \addlinespace[2pt]

    2023 & \textbf{NexusRaven}~\cite{Srinivasan2023NexusRaven} &
    Complex function calling &
    \xmark &
    \cmark &
    Curated functions \\

    2025 & \textbf{NESTFUL}~\cite{Basu2024NESTFUL} &
    Nested API calls &
    \cmark &
    \cmark &
    Curated APIs \\

    \bottomrule
  \end{tabularx}
\end{table}

\noindent\textbf{(1) Function-call validity.}
A first line of work evaluates whether a model can produce structurally correct
tool invocations once the relevant function is available. The Berkeley Function
Calling Leaderboard (BFCL)~\cite{Patil2025BFCL} has become a widely used
reference point for this setting, emphasizing function selection, argument
filling, and schema-constrained generation across diverse calling formats.
However, aggregate pass or fail scores often obscure where the failure actually
occurs. ToolEyes~\cite{Ye2025ToolEyes} therefore introduces a more fine-grained
diagnostic framework that separates errors in intent understanding, format
alignment, and tool selection. T-Eval~\cite{Chen2024TEval} pushes diagnosis
further by decomposing tool use into intermediate steps, enabling evaluation of
reasoning and execution decisions along the trajectory rather than only at the
final call.

\noindent\textbf{(2) Tool Retrieval and Selection.}
As the number of available tools grows, correctness is no longer limited to
schema compliance, but also depends on whether the model can retrieve and select
the right tool from a large candidate set. API-Bank~\cite{Li2023APIBank}
provides an early benchmark for this broader pipeline, covering API search,
planning, and response generation in simulated tool-use scenarios.
StableToolBench~\cite{Guo2024StableToolBench} extends this line toward more
reliable large-scale benchmarking, improving the reproducibility of evaluation
over real-world APIs. Gorilla~\cite{Patil2024Gorilla} complements these
benchmarks by stressing adaptation to evolving documentation, thereby testing
whether models can maintain correct tool use when API specifications change over
time.

\noindent\textbf{(3) Tool-Use Decisions.}
An equally important aspect of correctness is deciding whether tool use is
appropriate at all. Many earlier benchmarks implicitly assume that a tool should
be called, but realistic agents must also decide when to answer directly, when
to request clarification, and when to abstain. When2Call~\cite{Ross2025When2Call}
directly targets this decision boundary by evaluating whether models can choose
among tool invocation, follow-up questioning, and non-use. WTU-Eval~\cite{Ning2024WTUEval}
similarly studies whether-or-not tool usage across both tool-required and
tool-optional settings, highlighting that unnecessary invocation can degrade
overall performance even when the downstream tool is available.

\noindent\textbf{(4) Compositional Invocation.}
Finally, some benchmarks focus on correctness under compositional call
structures, where invocation quality depends on handling more complex signatures
or passing outputs across calls. NexusRaven~\cite{Srinivasan2023NexusRaven}
emphasizes zero-shot generalization from tool descriptions and remains
representative of benchmarking complex function-calling behavior without heavy
task-specific tuning. NESTFUL~\cite{Basu2024NESTFUL} sharpens this challenge by
evaluating nested sequences of API calls, where the output of one call becomes
the input of a subsequent one. These settings are still local in the sense that
they do not yet measure full end-to-end task completion, but they better capture
the compositional demands that modern agentic systems increasingly face.

Taken together, benchmarks in this layer show that tool usage correctness is a
multi-faceted notion. It includes not only syntactic validity, but also tool
selection under scale, calibrated decisions about when tool use is necessary,
and compositional accuracy in more complex invocation patterns. These
diagnostic benchmarks therefore provide the foundation for higher-level
evaluation, where the focus shifts from local invocation quality to complete
task execution and sustained interaction.

\subsection{Task Completion}
\label{subsec:task completion}

This category evaluates whether an agent can successfully complete an
end-to-end task when tools are used as functional components of the solution
pipeline, rather than merely producing a syntactically valid call. In contrast
to diagnostic correctness benchmarks, task-completion evaluations emphasize
final-task success under realistic inputs, domain constraints, and execution
feedback, thereby capturing both reasoning quality and tool-mediated
effectiveness. Representative benchmarks in this layer are summarized in
Table~\ref{tab:eval_task_completion}.

\begin{table}[!htbp]
  \centering
  \footnotesize
  \setlength{\tabcolsep}{2pt}
  \renewcommand{\arraystretch}{1.05}
  \caption{Representative benchmarks for task completion.}
  \label{tab:eval_task_completion}
  \rowcolors{2}{gray!10}{white}

  \begin{tabularx}{\columnwidth}{@{}
    >{\raggedright\arraybackslash}p{0.08\columnwidth}
    >{\raggedright\arraybackslash}p{0.25\columnwidth}
    >{\raggedright\arraybackslash}p{0.20\columnwidth}
    >{\raggedright\arraybackslash}p{0.18\columnwidth}
    >{\centering\arraybackslash}p{0.11\columnwidth}
    >{\centering\arraybackslash}p{0.12\columnwidth}
  @{}}
    \toprule
    \rowcolor{gray!15}
    \textbf{Year} &
    \textbf{Benchmark} &
    \textbf{Task Type} &
    \textbf{Tool Role} &
    \textbf{Executable} &
    \textbf{\shortstack{Open-Ended}} \\
    \midrule

    \rowcolor{white}
    \multicolumn{6}{@{}l@{}}{\textbf{Category I: Mathematical and Knowledge-Centric Reasoning Tasks}} \\

    2025 & \textbf{AIME 25}~\cite{MAA_AIME} &
    Competition math &
    Compute &
    \xmark &
    \xmark \\

    2023 & \textbf{TabMWP}~\cite{Lu2023TabMWP} &
    Table math reasoning &
    Compute &
    \xmark &
    \xmark \\

    2023 & \textbf{FreshQA}~\cite{Vu2024FreshLLMs} &
    Knowledge QA &
    Retrieve &
    \xmark &
    \xmark \\

    2023 & \textbf{ToolQA}~\cite{Zhuang2023ToolQA} &
    Tool-augmented QA &
    Retrieve + compute &
    \cmark &
    \xmark \\

    2021 & \textbf{MATH / GSM8K}~\cite{Hendrycks2021MATH,Cobbe2021GSM8K} &
    Foundation math &
    Compute &
    \xmark &
    \xmark \\

    2018 & \textbf{HotpotQA}~\cite{Yang2018HotpotQA} &
    Multi-hop QA &
    Retrieve &
    \xmark &
    \xmark \\

    \midrule

    \rowcolor{white}
    \multicolumn{6}{@{}l@{}}{\textbf{Category II: Programming and Data-Centric Task Execution}} \\

    2025 & \textbf{BigCodeBench}~\cite{Zhuo2024BigCodeBench} &
    Code generation &
    Execute &
    \cmark &
    \cmark \\

    2024 & \textbf{SWE-bench}~\cite{Jimenez2024SWEbench} &
    Software engineering &
    Execute + debug &
    \cmark &
    \cmark \\

    2024 & \textbf{BIRD / Spider 2.0}~\cite{Li2023BIRD,Lei2024Spider2} &
    Text-to-SQL &
    Query &
    \cmark &
    \cmark \\

    2024 & \textbf{DSBench}~\cite{Jing2024DSBench} &
    Data science &
    Execute &
    \cmark &
    \cmark \\

    2024 & \textbf{InterCode}~\cite{Yang2023InterCode} &
    Interactive coding &
    Execute + debug &
    \cmark &
    \cmark \\

    2024 & \textbf{SheetBench}~\cite{Ma2024SpreadsheetBench} &
    Spreadsheet tasks &
    Execute &
    \cmark &
    \cmark \\

    2021 & \textbf{HumanEval / MBPP}~\cite{Chen2021Codex,Austin2021ProgramSynthesisLLM} &
    Python coding &
    Execute &
    \cmark &
    \cmark \\

    \midrule

    \rowcolor{white}
    \multicolumn{6}{@{}l@{}}{\textbf{Category III: Scientific and Professional Domain Benchmarks}} \\

    2024 & \textbf{SciBench}~\cite{Wang2024SciBench} &
    Scientific reasoning &
    Compute &
    \xmark &
    \xmark \\

    2024 & \textbf{SciCode}~\cite{Tian2024SciCode} &
    Scientific coding &
    Execute &
    \cmark &
    \cmark \\

    2024 & \textbf{BioPlanner}~\cite{ODonoghue2023BioPlanner} &
    Biological protocols &
    Plan + retrieve &
    \xmark &
    \cmark \\

    2024 & \textbf{MedAgent-Bench}~\cite{Jiang2025MedAgentBench} &
    Clinical tasks &
    Plan + retrieve &
    \xmark &
    \cmark \\

    2023 & \textbf{ChemCrow}~\cite{Bran2024ChemCrow} &
    Chemistry workflows &
    Plan + execute &
    \cmark &
    \cmark \\

    \bottomrule
  \end{tabularx}
\end{table}

\noindent\textbf{(1) Mathematical and Knowledge-Centric Reasoning Tasks.}
A major line of task-completion evaluation studies whether agents can solve
reasoning problems more reliably when equipped with calculators, code
interpreters, search engines, or other external tools. MATH~\cite{Hendrycks2021MATH}
and GSM8K~\cite{Cobbe2021GSM8K} are widely reused as foundational testbeds for
measuring whether tool use improves multi-step numerical reasoning, while
AIME~\cite{MAA_AIME} and MATH-500~\cite{Lightman2024LetsVerify} further raise
the difficulty to competition-level mathematics. TabMWP~\cite{Lu2023TabMWP}
extends this setting to table-grounded mathematical reasoning, where solving the
task often requires parsing structured inputs and executing formula-based
computations. On the knowledge side, HotpotQA~\cite{Yang2018HotpotQA} remains a
representative benchmark for multi-hop retrieval and answer composition, and
ToolQA~\cite{Zhuang2023ToolQA} more directly evaluates end-to-end question
answering with external tools as functional components of the solution process.
FreshQA~\cite{Vu2024FreshLLMs} complements these benchmarks by focusing on
fast-changing world knowledge, thereby testing whether agents can complete
information-seeking tasks that static parametric knowledge alone cannot support.
More general QA benchmarks such as 2WikiMultihopQA~\cite{Ho2020TwoWikiMultihopQA},
Natural Questions~\cite{Kwiatkowski2019NaturalQuestions}, TriviaQA~\cite{Joshi2017TriviaQA},
MMLU~\cite{Hendrycks2021MMLU}, and StrategyQA~\cite{Geva2021StrategyQA} are also
frequently reused to study whether retrieval or tool augmentation improves final
answer quality.

\noindent\textbf{(2) Programming and Data-Centric Task Execution.}
A second line of evaluation focuses on whether agents can complete programming,
database, and data-analysis tasks under executable feedback. HumanEval~\cite{Chen2021Codex}
and MBPP~\cite{Austin2021ProgramSynthesisLLM} provide the classical starting
point for execution-based code generation, while BigCodeBench~\cite{Zhuo2024BigCodeBench}
substantially increases realism by requiring library-intensive code generation
across a large set of APIs. Beyond function-level synthesis, SWE-bench~\cite{Jimenez2024SWEbench}
evaluates repository-level software engineering by asking agents to resolve
real-world GitHub issues in existing codebases, making it especially relevant
for end-to-end agentic coding. BIRD~\cite{Li2023BIRD} and Spider 2.0~\cite{Lei2024Spider2}
extend task completion to enterprise-scale text-to-SQL settings, where success
depends on schema understanding, planning, and executable query generation.
DSBench~\cite{Jing2024DSBench} and DS-1000~\cite{Lai2023DS1000} further broaden
this space to realistic data-science workflows, from data cleaning to model
development. In more specialized settings, FinQA~\cite{Chen2021FinQA} targets
numerical reasoning over financial documents, SheetBench~\cite{Ma2024SpreadsheetBench}
addresses spreadsheet manipulation, and InterCode~\cite{Yang2023InterCode}
standardizes interactive coding with execution feedback, allowing agents to
debug and refine scripts during task solving.

\noindent\textbf{(3) Scientific and Professional Domain Benchmarks.}
Task-completion evaluation has also expanded into scientific and professional
domains, where tool use is required to satisfy domain-specific constraints
rather than only to improve generic reasoning. SciBench~\cite{Wang2024SciBench}
tests whether agents can solve physics and chemistry problems that often require
equation handling, unit-aware reasoning, and code-based computation. SciCode~\cite{Tian2024SciCode}
moves toward code-driven scientific problem solving in areas such as physics and
materials science. BioPlanner~\cite{ODonoghue2023BioPlanner} evaluates the
generation of biological protocols, emphasizing structured procedural planning
and precise function retrieval. ChemCrow~\cite{Bran2024ChemCrow} and
MedAgent-Bench~\cite{Jiang2025MedAgentBench} further demonstrate that
task-completion benchmarks now cover high-stakes workflows such as organic
synthesis and clinical diagnosis, where the agent must combine reasoning,
specialized tools, and domain knowledge to produce actionable solutions.

\subsection{Tool-Driven Interaction}
\label{subsec:tool-driven-interaction}

This category evaluates agents in dynamic and stateful environments where tool
use directly affects the external world. Unlike static QA or offline execution
benchmarks, these settings require agents to update plans as the environment
changes, recover from intermediate errors, and satisfy user goals under
realistic constraints. The focus therefore shifts from whether a tool call is
correct in isolation to whether a sequence of actions can reliably achieve a
goal in an interactive setting. Table~\ref{tab:eval_tool_driven_interaction}
summarizes representative benchmarks in this layer.

\begin{table}[!htbp]
  \centering
  \scriptsize
  \setlength{\tabcolsep}{2pt}
  \renewcommand{\arraystretch}{1.05}
  \caption{Evaluation benchmarks for tool-driven interaction.}
  \label{tab:eval_tool_driven_interaction}

  \rowcolors{2}{gray!10}{white}
  \begin{tabularx}{\columnwidth}{@{}
    >{\raggedright\arraybackslash}p{0.08\columnwidth}
    >{\raggedright\arraybackslash}p{0.26\columnwidth}
    >{\centering\arraybackslash}p{0.09\columnwidth}
    >{\raggedright\arraybackslash}p{0.20\columnwidth}
    >{\raggedright\arraybackslash}p{0.20\columnwidth}
    >{\raggedright\arraybackslash}X
  @{}}
    \toprule
    \rowcolor{gray!15}
    \textbf{Year} &
    \textbf{Benchmark} &
    \textbf{Trajectory} &
    \textbf{Target} &
    \textbf{Challenge} &
    \textbf{Signal} \\
    \midrule

    \rowcolor{white}
    \multicolumn{6}{@{}l@{}}{\textbf{Web Environments}} \\
    2024 & \textbf{WebArena}~\cite{Zhou2024WebArena} & \cmark & Web & Navigation & Task success \\
    2024 & \textbf{VisualWebArena}~\cite{Koh2024VisualWebArena} & \cmark & Web & Multimodal grounding & Task success \\
    2024 & \textbf{WebVoyager}~\cite{He2024WebVoyager} & \cmark & Web & Open-ended tasks & Task success \\
    2024 & \textbf{GAIA}~\cite{Mialon2024GAIA} & \cmark & General assistant & Multi-step tool use & Task success \\
    2024 & \textbf{TravelPlanner}~\cite{Xie2024TravelPlanner} & \cmark & Planning & Constraint satisfaction & Task success \\
    2023 & \textbf{WorldSense}~\cite{Benchekroun2023WorldSense} & \cmark & 3D environment & World-model reasoning & Task success \\
    2023 & \textbf{Mind2Web}~\cite{Deng2023Mind2Web} & \xmark & Web trajectories & Offline generalization & Trajectory match \\

    \midrule
    \rowcolor{white}
    \multicolumn{6}{@{}l@{}}{\textbf{Computer Environments}} \\
    2024 & \textbf{OSWorld}~\cite{Xie2024OSWorld} & \cmark & Desktop / OS & Long-horizon control & Task success \\
    2024 & \textbf{OS-Copilot}~\cite{Wu2024OSCopilot} & \cmark & Desktop / OS & Console and GUI control & Task success \\
    2024 & \textbf{OmniACT}~\cite{Kapoor2024OmniACT} & \cmark & Cross-app workflows & Scripted interaction & Task success \\
    2025 & \textbf{AndroidWorld}~\cite{Rawles2025AndroidWorld} & \cmark & Mobile / Android & App interaction & Task success \\
    2025 & \textbf{AppWorld / AppAgent}~\cite{Trivedi2024AppWorld,Zhang2025AppAgent} & \cmark & Mobile apps & App control & Task success \\
    2024 & \textbf{ScreenAgent}~\cite{Niu2024ScreenAgent} & \cmark & GUI & Visual UI automation & Task success \\
    2024 & \textbf{GUI-World}~\cite{Chen2024GUIWorld} & \cmark & GUI & Multimodal grounding & Task success \\
    2024 & \textbf{OfficeBench}~\cite{Wang2024OfficeBench} & \cmark & Office software & Procedural workflows & Task success \\
    2024 & \textbf{WorkArena}~\cite{Drouin2024WorkArena} & \cmark & Enterprise platform & Workflow completion & Task success \\
    2024 & \textbf{$\tau$-bench}~\cite{Yao2024TauBench} & \cmark & User + tools & Policy compliance & State outcome \\
    2024 & \textbf{TheAgentCompany}~\cite{Xu2024TheAgentCompany} & \cmark & Workplace simulation & Consequential tasks & Task success \\
    2024 & \textbf{CodeAct}~\cite{Wang2024CodeAct} & \cmark & Interactive environments & Code as action & Task success \\

    \midrule
    \rowcolor{white}
    \multicolumn{6}{@{}l@{}}{\textbf{Safety and Reliability}} \\
    2024 & \textbf{ToolSword}~\cite{Ye2024ToolSword} & \xmark & Tool outputs & Prompt-injection defense & Attack resistance \\
    2024 & \textbf{InjEcT-Agent}~\cite{Zhan2024InjecAgent} & \xmark & Tool outputs & Indirect prompt injection & Attack resistance \\
    2024 & \textbf{ToolEmu}~\cite{Ruan2024ToolEmu} & \cmark & Tool sandbox & Risk-aware interaction & Failure analysis \\
    2024 & \textbf{SafetyBench}~\cite{Zhang2024SafetyBench} & \xmark & Safety scenarios & Risk refusal & Safety judgment \\
    2024 & \textbf{R-Judge}~\cite{Yuan2024RJudge} & \xmark & Risk evaluation & Risk awareness & Risk judgment \\
    2024 & \textbf{PsySafe}~\cite{Zhang2024PsySafe} & \xmark & Psychological safety & Manipulation resistance & Safety judgment \\

    \bottomrule
  \end{tabularx}
\end{table}

\noindent\textbf{(1) Web Environments.}
WebArena~\cite{Zhou2024WebArena} and VisualWebArena~\cite{Koh2024VisualWebArena}
establish the core benchmark line for interactive web agents by providing
high-fidelity environments in which agents must navigate functional websites
through textual and visual observations. WebVoyager~\cite{He2024WebVoyager}
extends this setting to open-ended web tasks on real websites, requiring more
robust planning and multimodal grounding. Mind2Web~\cite{Deng2023Mind2Web}
complements these benchmarks with offline trajectories over diverse DOM
structures, making it useful for evaluating generalization before full online
deployment. Beyond browser navigation, GAIA~\cite{Mialon2024GAIA} evaluates
general assistant style tasks that require tool use, web access, and
multi-step problem solving, while TravelPlanner~\cite{Xie2024TravelPlanner}
focuses on constrained planning with external search tools. WorldSense~\cite{Benchekroun2023WorldSense}
further broadens interaction evaluation to grounded 3D environments, where the
agent must maintain a consistent world model while acting.

\noindent\textbf{(2) Computer Environments.}
A second line of work evaluates agents that interact with operating systems,
applications, and workplace software. OS-Copilot~\cite{Wu2024OSCopilot} and
OSWorld~\cite{Xie2024OSWorld} benchmark agents on desktop control through
terminal commands, GUI actions, and multimodal perception across realistic
computer tasks. OmniACT~\cite{Kapoor2024OmniACT}, ScreenAgent~\cite{Niu2024ScreenAgent},
and GUI-World~\cite{Chen2024GUIWorld} further emphasize grounded interaction
with graphical interfaces, while AndroidWorld~\cite{Rawles2025AndroidWorld},
AppWorld~\cite{Trivedi2024AppWorld}, and AppAgent~\cite{Zhang2025AppAgent}
extend this line to mobile environments. In workplace settings,
OfficeBench~\cite{Wang2024OfficeBench} and WorkArena~\cite{Drouin2024WorkArena}
evaluate interaction with enterprise platforms and office software, where task
success depends on long-horizon state tracking and procedural correctness.
Recent benchmarks move further toward realistic digital work. $\tau$-bench~\cite{Yao2024TauBench}
evaluates tool-agent-user interaction under domain policies and database state
constraints, while TheAgentCompany~\cite{Xu2024TheAgentCompany} benchmarks
agents on consequential workplace tasks that combine web browsing, code
execution, and collaboration in a simulated company environment. CodeAct~\cite{Wang2024CodeAct}
provides a complementary perspective by treating executable code as a unified
action space for interactive environments.

\noindent\textbf{(3) Safety and Reliability.}
As agents become more capable of acting in external environments, evaluation
must also examine whether they remain reliable and safe under adversarial or
high-risk conditions. ToolSword~\cite{Ye2024ToolSword} and InjEcT-Agent~\cite{Zhan2024InjecAgent}
focus on indirect prompt injection, testing whether malicious instructions
embedded in tool outputs can hijack subsequent agent behavior. ToolEmu~\cite{Ruan2024ToolEmu}
addresses this problem from a complementary angle by using an LLM-simulated
sandbox to emulate high-risk tool execution and expose dangerous failure modes
without causing real-world harm. R-Judge~\cite{Yuan2024RJudge} and
SafetyBench~\cite{Zhang2024SafetyBench} evaluate whether agents can identify
and refuse risky actions, while PsySafe~\cite{Zhang2024PsySafe} expands this
line to psychological safety and manipulation resistance. Together, these
benchmarks show that interaction-level evaluation is not only about completing
tasks, but also about maintaining reliability, policy compliance, and safe
behavior when actions have real consequences.

\section{Future Directions}
\label{sec:future-directions}

As the paradigm of agentic tool use matures from experimental prototypes to production grade
systems, we anticipate several critical trajectories that will define the next generation of
autonomous agents. These directions move beyond merely improving the accuracy of API calls, aiming
instead to reshape the fundamental architecture, interactivity, and societal integration of AI
agents.

{\renewcommand{\labelenumi}{(\arabic{enumi})}
\begin{enumerate}
  \item \textbf{Standardization and the MCP Revolution}

  The current ecosystem of tool-using agents is fragmented with bespoke integrations required for
  every new tool and framework. To overcome this scalability bottleneck the field is moving towards
  universal interoperability protocols most notably the Model Context Protocol (MCP)
  \cite{Hou2025MCP}. MCP serves as a standardized interface that decouples the agent from the
  specific implementation of tools allowing developers to build a tool once and have it be instantly
  usable by any compliant agent. This standardization transforms the integration problem from a
  quadratic complexity into a linear one. Beyond protocols we see a shift towards standardized
  orchestration patterns. OpenAI Swarm \cite{Rahman2025LLMSwarms} introduces ergonomic patterns for
  multi-agent orchestration exploring lightweight handoffs between specialized agents while AgentKit
  \cite{Wu2024AgentKit} provides a unified framework for "reinforcement fine-tuning" of agentic
  workflows.We foresee a future where tool libraries are not static lists
  but dynamic protocol-compliant marketplaces that agents can query and subscribe to in real-time.

  \item \textbf{Multimodal and Embodied Foundation Models}

  Future agents will transcend the limitations of text-based APIs to interact with the world through
  rich multimodal interfaces. Rather than relying solely on structured function calls agents are
  evolving to perceive and manipulate Graphical User Interfaces (GUIs) directly. A pivotal shift in
  2025 is the emergence of "Agentic Foundation Models" specifically pre-trained for action. Magma
  \cite{Yang2025Magma} represents this new breed a multimodal model capable of interpreting and
  grounding inputs across both digital and physical environments effectively bridging the gap between
  UI navigation and robotic manipulation. Complementing this ShowUI \cite{Lin2025ShowUI} introduces
  UI-guided visual token selection to handle high-resolution screens efficiently while OmniParser
  \cite{Lu2024OmniParser} provides a pure vision-based screen parsing module that extracts structured
  elements from screenshots significantly enhancing the action prediction capabilities of generalist
  models. Furthermore Agent-S \cite{Agashe2025AgentS} establishes an open agentic framework that
  uses computers like a human learning from experience to perform complex cross-application tasks.
  These advancements signal a move away from patching LLMs with vision encoders towards building
  native "Vision-Language-Action" models.

 \item \textbf{Agentic Self-Consistency and Continuous Evolution}

  To achieve true autonomy agents must move beyond reactive single-turn execution
  towards System 2 reasoning and lifetime learning. This entails the adoption of
  test-time scaling techniques where agents perform deep lookahead searches or
  self-consistency checks before executing critical tool calls. Frameworks like
  SearchR1 \cite{Jin2025SearchR1} have demonstrated the power of internalizing
  reasoning steps but future agents will go further by actively managing their
  memory. Titans \cite{Behrouz2025TitansMemorize} introduces a neural long-term
  memory module that learns to memorize historical context at test time enabling
  agents to handle "needle-in-a-haystack" retrieval tasks in massive interaction
  histories without retraining. System 2 Attention \cite{Weston2023System2Attention}
  enables agents to dynamically regenerate their own context to focus on relevant
  information while Trove \cite{Wang2024TroVE} facilitates the
  continuous evolution of an agent's tool library. Instead of resetting after each
  session future agents will curate their own internal libraries of synthesized
  tools and usage patterns achieving a form of self-consistency where their actions
  become increasingly efficient and aligned with their historical experience.Beyond improving tool use efficiency, reinforcement learning may also optimize behavior specific properties of agents. Character-R1 suggests that future agent optimization may extend to role aware reasoning and persona consistency in specialized settings \cite{Tang2026CharacterR1}.

  \item \textbf{Higher Safety Standards and Trustworthiness}

  As agents are granted the power to execute consequential actions such as
  modifying databases or managing financial transactions safety must be elevated
  from a secondary metric to a primary design constraint. Future research will
  focus on establishing bank-grade security protocols that go beyond simple
  alignment tuning. This includes developing robust defenses against indirect
  prompt injections that could hijack tool execution as highlighted by InjEcT-Agent
  \cite{Zhan2024InjecAgent}. Industrial solutions like NeMo Guardrails
  \cite{Rebedea2023NeMoGuardrails} are beginning to provide programmable safety
  layers that enforce behavioral boundaries while PromptArmor
  \cite{Shi2025PromptArmor} offers detection mechanisms for adversarial
  inputs. Crucially we must address the risk of deceptive alignment ensuring that
  agents do not learn to game reward systems by hiding errors or misrepresenting
  their tool usage a phenomenon investigated in Sleeper Agents
  \cite{Hubinger2024SleeperAgents}. Trustworthiness will rely on verifiable audit
  trails and sandboxed execution environments that guarantee human control even as
  agent autonomy increases.

  \item \textbf{Human-Agent Symbiosis and Workflow Reshaping}

  A final direction is human--agent symbiosis and workflow reshaping. As systems such as MetaGPT~\cite{Hong2023MetaGPT}, AutoGen~\cite{Wu2023AutoGen}, and OpenDevin~\cite{Wang2024OpenHands} show, the future of tool use may lie less in fully autonomous monolithic agents than in collaborative workflows that combine human oversight with specialized automated components. Relatedly, recent work on LLM based multi agent systems suggests that future agentic workflows may increasingly rely on system level coordination, role allocation, and orchestration across multiple specialized agents rather than a single monolithic agent \cite{Zhou2026ArchitectingMAS}.In this setting, skills may become a more practical abstraction than end-to-end autonomy: instead of requiring a single agent to solve every task from start to finish, systems can expose reusable capabilities such as retrieval, coding, spreadsheet editing, verification, or domain-specific analysis, and compose them flexibly under human supervision. This skill-centric view also helps explain why simpler and more structured systems can remain highly competitive. Agentless~\cite{Xia2025Agentless}, for example, suggests that carefully designed pipelines may outperform more complex long-horizon agents when task structure is clear. Together, these trends point toward a future in which agentic tool use is shaped by calibrated delegation, reusable skills, and workflow-aware collaboration between humans and agents.
\end{enumerate}
}

\section{Conclusion}
\label{sec:conclusion}

In this survey, we have systematically reviewed the technological landscape of agentic
tool use, examining the transition from prompting to dynamic autonomous optimization.
We categorized this domain into three distinct paradigms: Paradigm I leverages prompting
and context-based strategies to enable tool use in a plug-and-play fashion; Paradigm II
uses supervised learning to internalize robust tool-use capabilities; and Paradigm III
uses reward-driven policy learning to enable agents to adapt, self-correct, and recover
from complex execution errors. Alongside these paradigms, we treated evaluation as a
cross-cutting dimension and surveyed the evolving benchmark ecosystem, which has progressed from evaluating isolated
function call correctness to measuring holistic task success and safety in realistic
simulations. We hope this survey provides a coherent roadmap for researchers and
practitioners, and helps accelerate the development of tool-using agents towards being
more capable, reliable, and standardized autonomous agents.

\bibliographystyle{ACM-Reference-Format}
\bibliography{ref}

\appendix

\section{Summary of Related Works}
\label{appendix:summary_related_works}

Table~\ref{tab:summary_related_works} summarizes the major methodological papers discussed in this survey.

\begin{table*}[!htbp]
\caption{Summary of representative methodological papers on agentic tool use.}
\label{tab:summary_related_works}
\centering
\tiny
\setlength{\tabcolsep}{2.4pt}
\renewcommand{\arraystretch}{1.5}
\resizebox{\textwidth}{!}{
\rowcolors{1}{TableBg}{TableBg}
\begin{tabular}{C{2.2cm}c C{2.70cm} C{2.00cm} c c C{2.55cm} C{2.10cm}}
\toprule
\rowcolor{TableBg}
\textbf{Paper} & \textbf{Year} & \textbf{Training Signal} & \textbf{Tool Scope} & \textbf{Multi-turn} & \textbf{Multi-tool} & \textbf{Setting} & \textbf{Affiliation} \\
\midrule

ReAct~\cite{Yao2023ReAct} & 2023 & Prompting / ICL & External actions & \cmark & \cmark & Reason--act loop & Princeton \\ \midrule
Reflexion~\cite{Shinn2023Reflexion} & 2023 & Prompting + verbal feedback & External actions & \cmark & \cmark & Self-reflective interaction & Princeton \\ \midrule
CRITIC~\cite{Gou2024Critic} & 2024 & Prompting + interactive critique & Verification tools & \cmark & \cmark & Verification and correction & Tsinghua \\ \midrule
LATS~\cite{Zhou2023LATS} & 2023 & Prompting + search & Search & \cmark & \cmark & Trajectory tree search & UIUC \\ \midrule
ReWOO~\cite{Xu2023ReWOO} & 2023 & Prompting / planning & General external tools & \cmark & \cmark & Decoupled planning & MSU \\ \midrule
HuggingGPT~\cite{Shen2023HuggingGPT} & 2023 & Prompting / orchestration & Expert models & \cmark & \cmark & LLM tool orchestration & ZJU \\ \midrule
LLM-Compiler~\cite{Kim2024LLMCompiler} & 2024 & Prompting / planning & Function calls & \cmark & \cmark & Parallel decomposition & UC Berkeley \\ \midrule
PAL~\cite{Gao2023PAL} & 2023 & Prompting / code generation & Code execution & \xmark & \xmark & Program-aided reasoning & CMU \\ \midrule
PoT~\cite{Chen2023ProgramOfThoughts} & 2023 & Prompting / code generation & Code execution & \xmark & \xmark & Program-of-thought & Waterloo \\ \midrule
ViperGPT~\cite{Suris2023ViperGPT} & 2023 & Prompting / code generation & Vision APIs & \xmark & \cmark & Visual reasoning & Columbia \\ \midrule
Code as Policies~\cite{Liang2022CodeAsPolicies} & 2022 & Prompting / code generation & Control APIs & \xmark & \cmark & Embodied control & Google \\ \midrule
LATM~\cite{Cai2024LLMToolMakers} & 2024 & Prompting / tool creation & Python tools & \cmark & \cmark & Tool creation and reuse & Stanford \\ \midrule

Toolformer~\cite{Schick2023Toolformer} & 2023 & Self-supervised API annotation & APIs & \xmark & \cmark & API self-annotation & Meta AI \\ \midrule
AutoAct~\cite{Qiao2023AutoAct} & 2024 & Synthetic planning trajectories & Tool-augmented agents & \cmark & \cmark & Trajectory synthesis & ZJU \\ \midrule
ToolACE~\cite{Liu2024ToolACE} & 2025 & Self-evolving synthetic data & Function calls & \cmark & \cmark & API/dialogue synthesis & SJTU \\ \midrule
APIGen~\cite{Liu2024APIGen} & 2024 & Verifiable synthetic data & Function calls & \xmark & \cmark & API-call generation & Salesforce AI Research \\ \midrule
ToolLLM~\cite{Qin2023ToolLLM} & 2023 & Instruction tuning & RESTful APIs & \cmark & \cmark & Large-scale API use & Tsinghua \\ \midrule
API-Bank~\cite{Li2023APIBank} & 2023 & Benchmark-driven supervision & APIs & \cmark & \cmark & Planning and API use & Alibaba DAMO \\ \midrule
GPT4Tools~\cite{Yang2023GPT4Tools} & 2023 & Synthetic data + instruction & Tool collections & \xmark & \cmark & Tool-use tuning & Tsinghua SIGS \\ \midrule
Functionary~\cite{MeetKai2024Functionary} & 2024 & Instruction tuning & Function calls & \cmark & \cmark & Function calling & MeetKai \\ \midrule

FireAct~\cite{Chen2023FireAct} & 2023 & SFT on reasoning trajectories & General external tools & \cmark & \cmark & Reason--act tuning & System2 \\ \midrule
Agent-FLAN~\cite{Chen2024AgentFLAN} & 2024 & SFT with negative samples & Agent actions & \cmark & \cmark & Agent tuning & USTC \\ \midrule
ToolAlign~\cite{Chen2024ToolAlign} & 2024 & Alignment-oriented tuning & Tool-use environments & \cmark & \cmark & HHA alignment & RUC \\ \midrule
ToRA~\cite{Gou2024ToRA} & 2024 & Rationale-augmented training & Mathematical tools & \cmark & \cmark & Tool-augmented math & Tsinghua \\ \midrule
MetaTool~\cite{Huang2024MetaTool} & 2024 & Supervised tuning & Tool selection & \xmark & \cmark & Tool selection & XJTU \\ \midrule

ReTool~\cite{Feng2025ReTool} & 2025 & Reinforcement learning & General external tools & \xmark & \cmark & Strategic tool use & ByteDance Seed \\ \midrule
ToolExpander~\cite{Chen2025ToolExpander} & 2025 & RL + hard-sampling curriculum & General external tools & \cmark & \cmark & Capability expansion & OPPO \\ \midrule
ToolRL~\cite{Qian2025ToolRL} & 2025 & Outcome-reward RL & General external tools & \xmark & \cmark & Tool-use optimization & UIUC \\ \midrule
Agent0~\cite{Xia2025Agent0} & 2025 & RL + self-generated curriculum & General external tools & \cmark & \cmark & Self-evolving tool use & UNC \\ \midrule
Search-R1~\cite{Jin2025SearchR1} & 2025 & Reinforcement learning & Search / retrieval tools & \cmark & \xmark & Search-oriented reasoning & UIUC \\ \midrule
DeepResearcher~\cite{Zheng2025DeepResearcher} & 2025 & Reinforcement learning & Web research tools & \cmark & \cmark & Deep web research & SJTU \\ \midrule
AgentPRM~\cite{Xi2025AgentPRM} & 2025 & Process reward modeling & General external tools & \cmark & \cmark & Stepwise agent evaluation & FDU \\ \midrule
SimpleTIR~\cite{Xue2025SimpleTIR} & 2025 & End-to-end RL & General external tools & \cmark & \cmark & Joint reason--tool execution & NTU \\ \midrule
VerlTool~\cite{Jiang2025VerlTool} & 2025 & Agentic RL & Tool selector + memory & \cmark & \cmark & Holistic optimization & UWaterloo \\ \midrule
DeepAgent~\cite{Li2025DeepAgent} & 2025 & Agentic RL framework & Scalable toolsets & \cmark & \cmark & General tool agent & RUC \\ \midrule
ToolRM~\cite{Agarwal2025ToolRM} & 2025 & Reward modeling & Tool-use trajectories & \cmark & \cmark & Tool-call reward modeling & IBM Research \\ \midrule
Agent Q~\cite{Putta2024AgentQ} & 2024 & Guided MCTS + self-critique & Web environment & \cmark & \cmark & Interactive web agents & MultiOn \\ \midrule
DigiRL~\cite{Bai2024DigiRL} & 2024 & Offline-to-online RL & GUI / device control & \cmark & \cmark & Android interaction & UC Berkeley \\

\bottomrule
\end{tabular}
}
\end{table*}

\end{document}